\documentclass[11pt]{article}
\usepackage{acl}
\usepackage{times}
\usepackage{latexsym}
\usepackage{amsmath} 
\usepackage[T1]{fontenc}
\usepackage[utf8]{inputenc}
\usepackage{microtype}
\usepackage{inconsolata}

\usepackage{booktabs}
\usepackage{needspace}
\usepackage{enumitem}
\usepackage{float}

\usepackage{textcase}
\usepackage{graphicx}
\usepackage{adjustbox}
\usepackage{array}
\usepackage{orcidlink} 

\definecolor{acllink}{RGB}{0, 0, 150}
\hypersetup{
    colorlinks=true,
    linkcolor=acllink,
    citecolor=acllink,
    urlcolor=acllink
}

\usepackage{pgfplots}
\pgfplotsset{compat=1.17}
\usetikzlibrary{shapes.geometric,arrows.meta,positioning,calc,shadows.blur}

\definecolor{tc}{HTML}{2F6DB5}
\definecolor{ac}{HTML}{D9822B}
\definecolor{mc}{HTML}{6B4FA3}
\definecolor{gc}{HTML}{4A5A5A}

\newcolumntype{L}[1]{>{\raggedright\arraybackslash}p{#1}}
\usepackage{ragged2e}

\makeatother

\title{CamSpokenNLI: A Pilot Resource for Cross-Variety Transfer Evaluation in Cameroon Pidgin English and Camfranglais}

\author{
  \textbf{Tiani Pekins Ebika} \orcidlink{0009-0007-2550-3797} \\[1em]
  \begin{tabular}[t]{p{0.48\textwidth} p{0.48\textwidth}}
    \centering 
    Department of Computer Engineering \\
    Faculty of Engineering and Technology \\
    University of Buea \\
    Buea, Cameroon \\[2pt]
    \href{mailto:tiani.pekins@ubuea.cm}{ tiani.pekins@ubuea.cm}
    & 
    \centering 
    LAB  237 \\
    Cameroon Open Research Community \\
    \href{https://lab237.org}{lab237.org} \\
    Buea, Cameroon \\[2pt]
    \href{mailto:tiani@lab237.org}{ tiani@lab237.org}
  \end{tabular}
}

\begin{document}
\maketitle

\begin{abstract}
Cameroon Pidgin English and Camfranglais are widely used in Cameroon, yet to our knowledge, neither has a dedicated spoken Natural Language Inference (NLI) resource. We introduce CamSpokenNLI, a pilot dataset of 114 scenario-matched spoken NLI items in each variety (228 in total), recorded by one native speaker from Buea. Each item pairs premise audio and text with a hypothesis, an NLI label, and English glosses. The author assigned labels, so the dataset is an exploratory resource rather than a mature benchmark.
The NLI experiments use the reference text; the audio is evaluated separately. XLM-RoBERTa baselines with five-fold item-level cross-validation reach 44.7\% on Pidgin and 46.2\% on Camfranglais (chance 33.3\%). Still, neither is statistically distinguishable from hypothesis-only input, so hypothesis wording may account for much of the accuracy. Under disjoint-item cross-variety transfer, where a test item's counterpart is never seen in training, accuracy is 40.1\% (Pidgin to Camfranglais) and 45.6\% (Camfranglais to Pidgin). A counterpart-leak audit raises this to 60--64\%, illustrating how counterpart leakage can overstate transfer.
Zero-shot Whisper transcription of the recordings performs poorly: the best word error rates are 66.6\% for Pidgin and 45.5\% for Camfranglais, and with automatic language identification, Whisper Large-v3 labels 93 of 114 Pidgin recordings as Yoruba or Swahili.
Together, the results suggest that such resources need disjoint-item evaluation and hypothesis-only controls, and that speech recognition must be adapted to these varieties before spoken NLI can be tested end to end.
\end{abstract}

\section{Introduction}

In Cameroon, Pidgin English (Kamtok, Wes Cos; ISO 639-3: \textit{wes}) and Camfranglais, which is a French--English--Pidgin contact variety shaped by local languages, are mainly spoken rather than written
\citep{echu2006,ozon2017spoken}. A client who needs an electrician is more likely to say ``\textit{light don go fo haus}'' (``the power has gone off at home'') than to type it out. Yet \mbox{systems} capable of reasoning over such utterances remain limited, particularly for Natural Language Inference (NLI), which determines whether a statement entails, contradicts, or is neutral with respect to another statement.

A large speaker population does not guarantee adequate computational coverage. Cameroon Pidgin is spoken in some form by an estimated half of Cameroon's population \citep{lewis2014}. \mbox{However}, its speakers are hard to count because competence ranges from native use to a trade \mbox{language} \citep{schroder2013}. It belongs to the wider West African Pidgin English (WAPE) \mbox{cluster}, which \citet{yakpo2024} estimates reaches up to 140 million speakers across six nations (Nigeria, Cameroon, Ghana, Sierra Leone, Equatorial Guinea, and The Gambia). Within this cluster, Nigerian Pidgin has benefited from growing computational resources for tasks such as sentiment analysis \citep{muhammad2022} and translated NLI data \citep{kondoro2026}.
Still, resources for a related regional variety cannot replace direct evaluation of Cameroon Pidgin, Camfranglais, and related Cameroonian contact varieties.

Large English NLI datasets such as SNLI \citep{bowman2015} and MultiNLI \citep{williams2018} exist, and recent efforts have extended NLI-style resources to African languages through translation of XNLI \citep{adelani2025,kondoro2026}. Audio-premise entailment has also begun to receive attention, for non-speech audio \mbox{\citep{deshmukh2025}} and for African-accented English \citep{okocha2026}. Despite these developments, our review of the published literature and publicly available resources did not identify a dedicated NLI dataset for Cameroon Pidgin English or Camfranglais, nor a spoken-premise NLI resource for either variety.
To address this gap, we introduce CamSpokenNLI, a spoken-premise NLI resource for Cameroon Pidgin and Camfranglais, and evaluate it with text-based baselines and zero-shot speech recognition.

\subsection{Motivation}

The need is practical as well as scientific, especially for voice-based applications. CamSpokenNLI was developed alongside LocalHands Africa, a voice-first service-exchange platform in which a spoken request such as ``\textit{light don go fo haus}'' must be matched to an artisan skill like electrical work, whichever variety the speaker uses. Entailment offers one way to test such matching: a system should be able to judge that ``The speaker needs an electrician'' can reasonably be inferred from the spoken request. We disclose this connection \mbox{explicitly}; see the \hyperref[sec:ethics]{\textit{Ethics Statement}}.

\subsection{Contributions}
The contributions of this work are as follows:

\vspace{10pt}
\noindent\textbf{(1)} We introduce CamSpokenNLI, a pilot spoken-premise NLI resource containing 114 manually annotated, sce\-nario-matched items in each variety. To the best of our knowledge, it is the first dedicated NLI resource for either variety and the first with spoken premises in these varieties.

\vspace{10pt}
\noindent\textbf{(2)} We report XLM-RoBERTa baselines under five-fold item-level cross-validation, together with a hypothesis-only analysis. The XLM-RoBERTa accuracy is not statistically distinguishable from the hypothesis-only baselines, so hypothesis wording may account for much of it.

\vspace{10pt}
\noindent\textbf{(3)} We provide a cross-variety transfer study on disjoint items, and an audit illustrating how \mbox{counterpart} leakage can inflate transfer accuracy.

\vspace{10pt}
\noindent\textbf{(4)} We assess audio-grounded robustness for both varieties by transcribing the premise recordings with zero-shot Whisper (Base, Small, and \mbox{Large-v3}) under automatic, forced-English, and forced-French decoding, analyzing failure patterns such as language misidentification and lexical substitution. This evaluation is separate from the NLI baselines, which use the reference text only.

\vspace{10pt}
\noindent\textbf{(5)} We release the recording protocol, the complete dataset, and reproducible Colab notebooks so that every NLI run, prediction, and Whisper transcript can be inspected by the community.

\vspace{10pt}
\noindent CamSpokenNLI is too small for a standard benchmark; we frame it as a seed resource whose protocol and baselines can guide larger-scale collection by the broader African NLP community.

\section{Related Work}
\label{sec:related_work}
We organize prior work by relevance to spoken Cameroonian NLI, then distinguish the closely related Nigerian Pidgin resources from Cameroon Pidgin English and Camfranglais.
\subsection{Prior Work in NLI and African NLP}
\label{subsec:prior_work_nli}

\begin{itemize}

    \item \textbf{Foundational English NLI:} SNLI \citep{bowman2015} and MultiNLI \citep{williams2018} established large-scale NLI resources for English. XNLI \citep{conneau2018} subsequently extended multilingual NLI evaluation to English and 14 other languages.

    \item \textbf{Annotation artifacts:} On English NLI datasets, classifiers that see only the hypothesis can score well above the majority-class baseline, which indicates that some labels can be predicted from hypothesis wording alone \citep{gururangan2018,poliak2018}. This motivates the hypothesis-only and premise-only diagnostics we run.
    
   \item \textbf{Multilingual NLI:} XLM-RoBERTa \citep{conneau2020}, pretrained on CommonCrawl text in 100 languages, provides a widely used multilingual encoder for low-resource transfer settings; Cameroon Pidgin English and Camfranglais are not identified as dedicated pretraining languages.
    
   \item \textbf{African NLU resources:} Community-driven efforts have expanded computational resources for African languages, including a Twitter sentiment corpus that covers Nigerian Pidgin \citep{muhammad2022}.
    
   \item \textbf{NLI for African languages:} IrokoBench \citep{adelani2025} includes AfriXNLI, a human translation of XNLI into African languages; Afri-XNLI+ \citep{kondoro2026} adds further translated XNLI pairs, including Nigerian Pidgin. These resources are \mbox{text-only} translations of English items. Our review did not identify an NLI resource for Cameroon Pidgin English or Camfranglais, nor one built from original, \mbox{spoken Cameroonian material}.
    
   \item \textbf{Speech and entailment:} A spoken corpus of Cameroon Pidgin English \citep{ozon2017spoken} provides recordings and transcripts, including data from Buea, for linguistic study; however, it lacks inference labels. In the domain of reasoning, Audio Entailment \citep{deshmukh2025} pairs general (non-speech) audio premises with text hypotheses, while Afrispeech Semantics \citep{okocha2026} evaluates entailment on African-accented English. To our knowledge, these entailment studies do not cover Cameroon Pidgin English or Camfranglais.
    
   \item \textbf{Sociolinguistic research:} Both varieties are well documented, including work on Cameroon Pidgin English \citep{ayafor2017,ozon2017spoken} and on the structure and use of Camfranglais \citep{kouega2003,ubanako2021}. The gap \mbox{addressed} here is therefore computational rather than descriptive, particularly for inference tasks such as NLI over spoken premises.
    
   \item \textbf{Creole-language NLI precedent:} Outside Africa, JamPatoisNLI \citep{armstrong2022} provides a precedent for constructing NLI \mbox{resources} for English-lexified creole languages. Comparable resources for African pidgin and contact varieties remain scarce, especially spoken Cameroonian NLI data.
\end{itemize}
\vspace{-4pt}
\noindent Translated benchmarks make results comparable across languages, but a translation inherits the wording of its English source and may miss locally used expressions. For contact varieties, whose vocabulary and constructions mix several source languages, items written directly in the variety can differ from translated ones. CamSpokenNLI takes this route: premises and hypotheses are written in each variety, and the premises are recorded.

\subsection{Distinctions from Nigerian Pidgin Resources}
\label{subsec:nigerian_pidgin_distinctions}
Existing Nigerian Pidgin resources (such as \mbox{NaijaSenti} and Afri-XNLI+) cannot stand in for Cameroon Pidgin English. The two Pidgins belong to the same WAPE cluster \citep{yakpo2024}, but they differ in ways that matter for speech and NLI data.

\begin{itemize}
 \item \textbf{Contact languages:} As contact languages, both Pidgins are English-lexified, yet they \mbox{diverge} in their substrate influences. While Nigerian Pidgin primarily incorporates \mbox{Benue-Kwa} languages of the Niger-Congo family \citep{faraclas2013}, Cameroon Pidgin English draws significantly on French and \mbox{indigenous} Cameroonian languages \citep{schroder2013}. Earlier estimates reported by \citet{echu2003} put the Cameroon Pidgin English lexicon at roughly 80--85\% English and 13--14\% indigenous loanwords, with French contribution rising to approximately 2--5\%.

  \item \textbf{Grammar:} The imperfective/progressive marker is typically \textit{di} in Cameroon Pidgin English \citep{schroder2013}, whereas Nigerian Pidgin uses \textit{dè} (or \textit{dey}) \citep{faraclas2013}. \mbox{Examples} include ``\textit{How you di do?}'' (Cameroon) and ``\textit{How you dey?}'' (Nigeria).
 
\item \textbf{Camfranglais:} Unlike Nigerian Pidgin, Camfranglais is a distinct mixed variety rather than an English-lexified pidgin, and it is also separate from Cameroon Pidgin English. Its lexicon comes from French, English, West African Pidgin, Cameroonian languages, Latin, and Spanish \citep{kouega2003}. To date, no existing Nigerian Pidgin resource covers it, so it needs its own evaluation data.
\end{itemize}
\vspace{-4pt}
\noindent These differences motivate treating Cameroon Pidgin English and Camfranglais as distinct computational targets; conflating them would obscure the unique Francophone and indigenous lexical integration central to the Cameroonian context. We \mbox{reference} Nigerian Pidgin to situate our work within the broader landscape of West African \mbox{creoles}, serving as linguistic context rather than an experimental baseline. Consequently, Section \ref{subsec:results} evaluates cross-variety transfer directly between the two Cameroonian varieties.

\section{Dataset Construction}
This section describes how the first release (v1.0) was constructed.

\subsection{Design}
Each sample is a tuple:
\begin{equation*}
\small
\begin{gathered}
(\texttt{id}, \texttt{premise\_text}, \texttt{premise\_text\_en}, \texttt{premise\_audio}, \\
\texttt{hypothesis\_text}, \texttt{hypothesis\_text\_en}, \texttt{label})
\end{gathered}
\end{equation*}
\noindent where \texttt{label} is one of the three NLI classes (\texttt{entailment}, \texttt{contradiction}, or \texttt{neutral}). The \texttt{\_en} fields are English glosses of the premise and hypothesis, included so that readers who do not speak Cameroon Pidgin English or Camfranglais can inspect the data; they are translations of the original material, not independently authored English NLI examples. Labels are balanced within each variety: 38 entailment / 38 contradiction / 38 neutral. Labels follow a fixed rule: a hypothesis is entailment if it must be true whenever the premise is true, contradiction if the two cannot both be true, and neutral otherwise. All premises are declarative sentences; questions and imperatives are not used.

The dataset is organized as two variety subsets that follow the same schema. They are scenario-matched: the same 114 scenarios are represented in both varieties, with each variety providing its own sentences and recording, and corresponding items sharing the same numeric item identifier (\texttt{wes\_NNNNN} / \texttt{camf\_NNNNN}). Labels agree across all 114 corresponding items, and the English premise and hypothesis glosses are identical across the two files. Each subset keeps its own annotations and \mbox{audio}, so either variety can be inspected or \mbox{extended} independently.
All 114 premises and all 114 hypotheses are unique within each variety.

\subsection{Sentence Creation}
\label{sec:sentence}
Sentences reflect daily life in Buea/Molyko and cover the everyday service domains where voice-first requests arise, such as electricity, water, transport (bike/moto), markets, school, and financial transactions. They were drawn from the author's everyday experience in these settings and resemble everyday exchanges, for example, with friends or a motorbike rider. Each premise was recorded in the target variety with a hypothesis already in mind that stood in a specific relation to it \mbox{(entailment}, contradiction, or neutral), and the label was fixed at that point. The premise recordings were then transcribed, the \mbox{hypotheses} were written as text, and English glosses were added afterward as translations of both. Only premises are recorded in the release; hypotheses are text only. The author wrote all premises and hypotheses for this dataset. Premises are short \mbox{single} sentences averaging 9.2 words (Pidgin) and 9.5 (Camfranglais); hypotheses average 6.9 and 7.4 words, respectively, and vary little by label (Appendix~\ref{app:dataset_checks}).

The Pidgin items were produced first. For Camfranglais, the author expressed the same 114 scenarios in his everyday Camfranglais, with his own sentences and recordings, transcribed the same way and matched to the same English gloss. The \mbox{Camfranglais} sentences are therefore not word-for-word translations of the Pidgin ones: the mean word overlap between counterpart premises, measured as the Jaccard similarity of their word sets, is about 8\%, and no pair is identical (Appendix~\ref{app:dataset_checks}). Because each pair of items describes the same \mbox{situation} under the same label, the two subsets are scenario-matched rather than independent. A test item's counterpart in the other variety shares its meaning and label, so cross-variety transfer is \mbox{evaluated} with disjoint-item splits (Section~\ref{subsec:setup}).

\medskip
\Needspace{16\baselineskip}
\noindent\textbf{Example (Cameroon Pidgin English):}
\begin{center}
\small
\setlength{\extrarowheight}{3pt}
\renewcommand{\arraystretch}{1.2}
\begin{tabular}{L{2.1cm} L{2.1cm} c}
\toprule
\textbf{Premise / Gloss} & \textbf{Hypothesis / Gloss} & \textbf{Label} \\
\midrule
light no de molyko fo two days na so. \par\vspace{2pt} \emph{(The power has been off in Molyko for two days.)} & molyko no get light. \par\vspace{2pt} \emph{(Molyko doesn't have electricity.)} & entailment \\
\hline
papa de fo haus di wait you. \par\vspace{2pt} \emph{(Dad is home right now waiting for you.)} & papa de fo workplace. \par\vspace{2pt} \emph{(Dad is at work.)} & contradiction \\
\hline
a jus take salary na na so. \par\vspace{2pt} \emph{(I’ve just received my salary.)} & a go pay school fees wit de moni. \par\vspace{2pt} \emph{(I'll pay school fees with the money.)} & neutral \\
\bottomrule
\end{tabular}
\end{center}

\medskip
\noindent\textbf{Example (Camfranglais):}\nopagebreak
\begin{center}
\small
\setlength{\extrarowheight}{3pt}
\renewcommand{\arraystretch}{1.2}
\begin{tabular}{L{2.1cm} L{2.1cm} c}
\toprule
\textbf{Premise / Gloss} & \textbf{Hypothesis / Gloss} & \textbf{Label} \\
\midrule
la light est off à Molyko depuis deux jours. \par\vspace{2pt} \emph{(The power has been off in Molyko for two days.)} & Molyko n'a pas la light. \par\vspace{2pt} \emph{(Molyko doesn't have electricity.)} & entailment \\
\hline
le vieux est au piol entrain de te wait. \par\vspace{2pt} \emph{(Dad is home right now waiting for you.)} & le vieux est au work. \par\vspace{2pt} \emph{(Dad is at work.)} & contradiction \\
\hline
mon salaire vient de tomber. \par\vspace{2pt} \emph{(I've just received my salary.)} & je vais use les doo là pour pay les school fees. \par\vspace{2pt} \emph{(I'll pay school fees with the money.)} & neutral \\
\bottomrule
\end{tabular} 
\end{center}

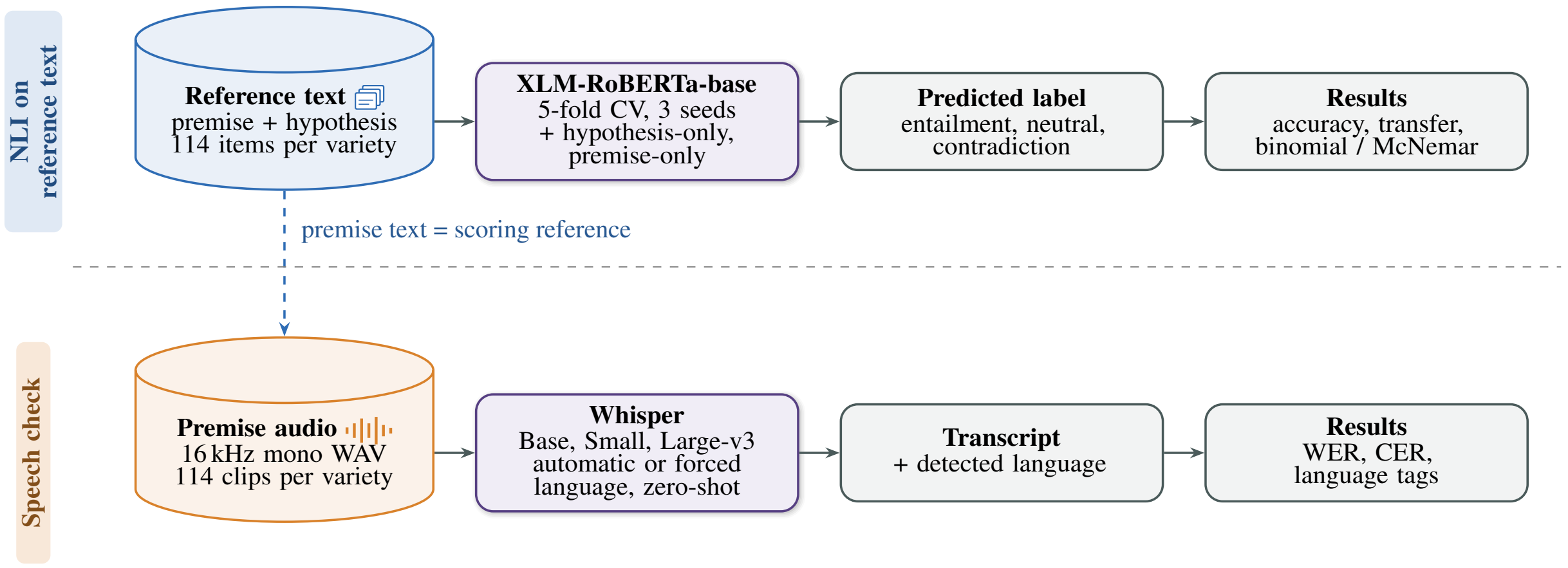
\begin{figure*}[t]
\centering
\resizebox{\textwidth}{!}{%
\begin{tikzpicture}[
  font=\footnotesize,
  base/.style={align=center, inner sep=4pt, text width=3.7cm, minimum height=1.2cm},
  data/.style={base, cylinder, shape border rotate=90, aspect=0.22, draw, thick, minimum height=1.75cm, text width=3.4cm},
  mod/.style={base, draw, thick, rounded corners=6pt, fill=mc!10, draw=mc!80!black,
    blur shadow={shadow xshift=0.6pt, shadow yshift=-0.6pt, shadow blur steps=4, shadow opacity=35}},
  res/.style={base, draw, rounded corners=6pt, fill=gc!7, draw=gc, thick},
  arr/.style={-{Stealth[length=5pt]}, thick, draw=gc},
  band/.style={rotate=90, font=\footnotesize\bfseries, align=center, text width=2.6cm, inner sep=2pt}
]
\node[data, draw=tc, fill=tc!10] (t1) at (0,0) {\textbf{Reference text}~\tikz[baseline=-0.5ex]{\foreach \i in {2,1,0}{\draw[tc,fill=white,line width=0.5pt,rounded corners=0.5pt](\i*0.045,\i*0.045-0.13) rectangle ++(0.26,0.2);}\draw[tc,line width=0.4pt](0.04,-0.02)--(0.22,-0.02);\draw[tc,line width=0.4pt](0.04,-0.07)--(0.17,-0.07);}\\[-1pt]\footnotesize premise + hypothesis\\[-2pt]\footnotesize 114 items per variety};
\node[mod, right=0.5cm of t1] (t2) {\textbf{XLM-RoBERTa-base}\\[-1pt]\footnotesize 5-fold CV, 3 seeds\\[-2pt]\footnotesize + hypothesis-only,\\[-2pt]\footnotesize premise-only};
\node[res, right=0.5cm of t2] (t3) {\textbf{Predicted label}\\[-1pt]\footnotesize entailment, neutral,\\[-2pt]\footnotesize contradiction};
\node[res, right=0.5cm of t3] (t4) {\textbf{Results}\\[-1pt]\footnotesize accuracy, transfer,\\[-2pt]\footnotesize binomial / McNemar};
\foreach \a/\b in {t1/t2,t2/t3,t3/t4}{\draw[arr] (\a) -- (\b);}
\node[data, draw=ac, fill=ac!10, below=1.8cm of t1] (a1) {\textbf{Premise audio}~\tikz[baseline=-0.5ex]{\foreach \x/\h in {0/0.08,0.09/0.18,0.18/0.3,0.27/0.18,0.36/0.34,0.45/0.14,0.54/0.07}{\draw[ac,line width=1pt](\x,-\h/2)--(\x,\h/2);}}\\[-1pt]\footnotesize 16\,kHz mono WAV\\[-2pt]\footnotesize 114 clips per variety};
\node[mod, right=0.5cm of a1] (a2) {\textbf{Whisper}\\[-1pt]\footnotesize Base, Small, Large-v3\\[-2pt]\footnotesize automatic or forced\\[-2pt]\footnotesize language, zero-shot};
\node[res, right=0.5cm of a2] (a3) {\textbf{Transcript}\\[-1pt]\footnotesize + detected language};
\node[res, right=0.5cm of a3] (a4) {\textbf{Results}\\[-1pt]\footnotesize WER, CER,\\[-2pt]\footnotesize language tags};
\foreach \a/\b in {a1/a2,a2/a3,a3/a4}{\draw[arr] (\a) -- (\b);}
\draw[arr, dashed, draw=tc] (t1.south) -- node[pos=0.28, right=2pt, font=\footnotesize, align=left, text=tc!80!black] {premise text = scoring reference} (a1.north);
\coordinate (dl) at ($(t1.south west)!0.5!(a1.north west)+(-1.55,0)$);
\draw[dashed, gray] (dl) -- ([xshift=0.4cm]dl -| t4.east);
\node[band, fill=tc!15, text=tc!70!black, rounded corners=3pt] at ($(t1.west)+(-1.25,0)$) {NLI on\\reference text};
\node[band, fill=ac!18, text=ac!60!black, rounded corners=3pt] at ($(a1.west)+(-1.25,0)$) {Speech check};
\end{tikzpicture}%
}
\caption{Overview of the two separate paths in this paper. Top: NLI is run on the reference text of the premises and hypotheses with XLM-RoBERTa-base. Bottom: the premise audio is transcribed by zero-shot Whisper and scored against the premise text. The results are in Sections~\ref{subsec:results} (NLI) and~\ref{sec:asr} (speech check). Neither the audio nor the ASR transcripts are used as NLI input.}
\label{fig:pipeline}
\end{figure*}

\subsection{Audio Protocol and Pipeline Overview}
\label{sec:audio}
Each premise was recorded as one spoken sentence. Figure~\ref{fig:pipeline} shows how the data are used: NLI classification with XLM-RoBERTa-base takes the premise and hypothesis text as input, while the premise audio is used only in a separate speech check, where zero-shot Whisper transcribes it, and the transcript is scored against the premise text (Section~\ref{sec:asr}). The recording protocol was as follows:

\begin{itemize}[itemsep=-2pt, topsep=2pt]
    \item \textbf{Format:} WAV, 16 kHz, mono, 16-bit PCM
    \item \textbf{Length:} 2.0--6.8\,s per clip after trimming (mean 3.7\,s for Pidgin, 3.4\,s for Camfranglais), one sentence per file
    \item Natural conversational pace (not read-aloud/news style), consistent volume
    
    \item 0.1\,s of silence before and 1.0\,s of silence after the speech
    \item Quiet room, phone $\sim$15 cm from mouth, \mbox{minimal} background noise
    \item Conversion with \texttt{ffmpeg} in a Colab notebook (.m4a $\rightarrow$ .wav, 16 kHz,     mono)
\end{itemize}

\subsection{Splits and Class Balance}
\label{sec:splits_balance}
The dataset contains 114 examples per language variety and is released with a single training split. No dedicated development or held-out test partition is included. This design reflects the early stage of the release: the current objective is to establish a reproducible resource and baseline rather than a mature benchmark for held-out evaluation. Each language subset is balanced across the three NLI classes \mbox{(38 entailment, 38 contradiction, 38 neutral)}.

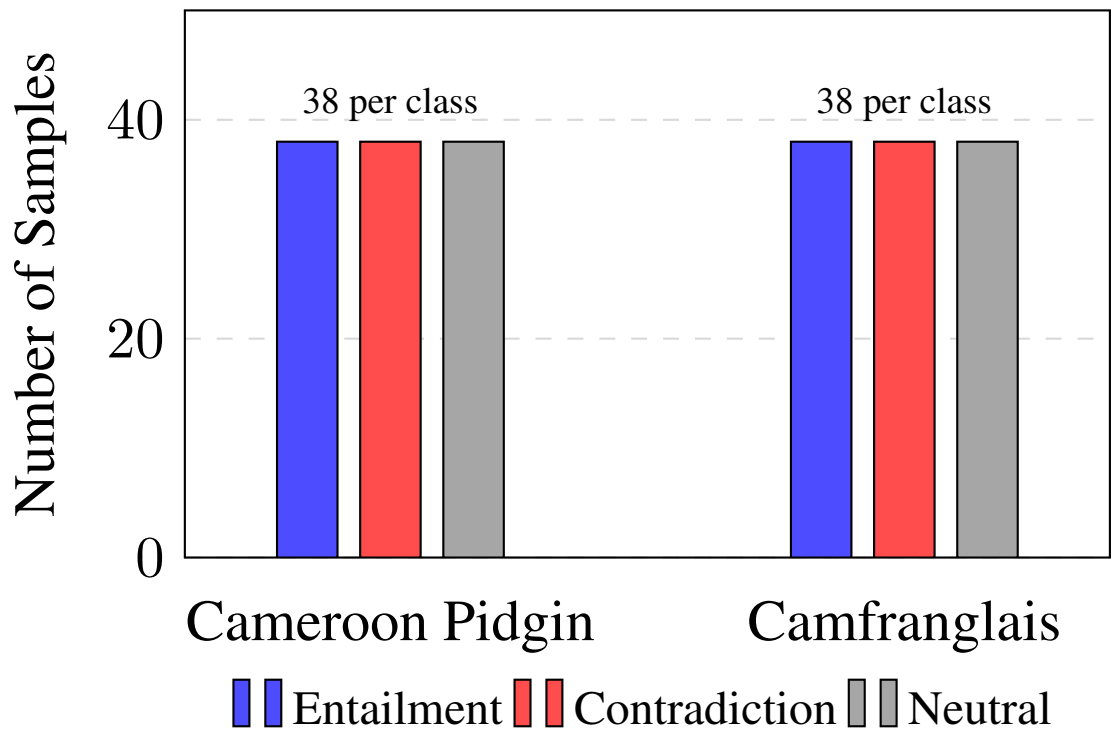
\begin{figure}[H]
    \centering
    \begin{tikzpicture}
    \begin{axis}[
        ybar=0.15cm,
        bar width=0.4cm,
        ylabel={Number of Samples},
        symbolic x coords={Cameroon Pidgin, Camfranglais},
        xtick=data,
        enlarge x limits=0.4,
        ymin=0, ymax=50,
        ymajorgrids=true,
        grid style={dashed,gray!30},
        legend style={at={(0.5,-0.2)}, anchor=north, legend columns=-1, draw=none, font=\footnotesize},
        legend image code/.code={\draw[#1] (-1.5pt,-0.25em) rectangle (1.5pt,0.55em); \draw[#1] (4.5pt,-0.25em) rectangle (7.5pt,0.55em);},
        axis line style={-},
        major tick style={draw=none},
        width=\columnwidth,
        height=5.2cm
    ]

    \addplot[fill=blue!70, draw=black] coordinates {(Cameroon Pidgin, 38) (Camfranglais, 38)};
    \addplot[fill=red!70, draw=black] coordinates {(Cameroon Pidgin, 38) (Camfranglais, 38)};
    \addplot[fill=gray!70, draw=black] coordinates {(Cameroon Pidgin, 38) (Camfranglais, 38)};

    \node[above, font=\scriptsize] at (axis cs:Cameroon Pidgin,38) {38 per class};
    \node[above, font=\scriptsize] at (axis cs:Camfranglais,38) {38 per class};

    \legend{Entailment, Contradiction, Neutral}
    \end{axis}
    \end{tikzpicture}
    \caption{Label distribution in each variety subset: 114 samples, balanced across entailment, contradiction, and neutral (38 each). Chance level: 33.3\%.}
    
    \label{fig:balance}
\end{figure}

\noindent {Predicting one label for every item is correct for exactly one third (38 of 114), the chance level in the exact binomial tests. For the experiments, the 114 items are divided once into five stratified folds \mbox{(22--23 items per fold, 7--8 per class)}. Both varieties share the same fold assignment: item $i$ is in the same fold in Pidgin and in Camfranglais. It is also fixed across all random seeds and all conditions, so seeds change only model initialization and training order. In each run, one fold is held out for testing, and the other four are used for training, so across the five runs every item is held out exactly once. In cross-variety transfer, the model is trained on one variety’s items outside the held-out fold and tested on the other variety’s items inside it, so a test item and its counterpart are never in the training set. The folds were generated once, with seed 42, and were never redrawn.  The full fold item-level assignment is given in Appendix~\ref{app:folds}}.

\subsection{Positionality Statement}
\label{sec:positionality}
All recordings are by a single speaker (author, male, native speaker of both varieties, based in Buea). This limits regional and demographic \mbox{coverage}. It is also a strength: sentences and labels reflect the author's firsthand native-speaker judgment and lived experience of both varieties, rather than data elicited from third parties. The same person wrote, recorded, and labeled every item, so no \mbox{inter-annotator} agreement is available. The two subsets are scenario-matched and share the same 114 scenarios, so they are not fully independent.

\section{Experiments and Results}
\label{sec:experiments}
We run two separate evaluations (Figure~\ref{fig:pipeline}): NLI on the reference text of the premises and hypotheses (Sections~\ref{subsec:setup}--\ref{subsec:results}), and a zero-shot Whisper check on the premise audio (Section~\ref{sec:asr}). Each gives its setup and then its results; we then interpret the two together in Section~\ref{subsec:interpretation}.

\subsection{NLI Experimental Setup}
\label{subsec:setup}

\noindent\textbf{Approach.} We report a text-based NLI baseline using XLM-RoBERTa \citep{conneau2020} on the reference text of the premises and hypotheses. The task is a three-way classification: entailment, contradiction, or neutral. Because the dataset contains only 114 examples per variety and has no dedicated development or test partition, we evaluate with stratified five-fold cross-validation over items rather than a held-out test set.

\medskip
\noindent\textbf{Model and training.} We fine-tune \texttt{xlm-\allowbreak roberta-\allowbreak base} with the following configuration (full details in Appendix~\ref{app:config}):

\begin{itemize}[itemsep=0pt, topsep=6pt]
    \item 15 training epochs, batch size 4, learning rate $2 \times 10^{-5}$;
    \item maximum sequence length 64;
    \item random seeds 42, 123, and 2024.
\end{itemize}
Training used AdamW with a linear learning-rate schedule, weight decay 0.0, and fp16 mixed precision, with an evaluation batch size of 8, on a single NVIDIA Tesla T4 GPU in Google Colab.

\medskip
\noindent\textbf{Epochs and tuning.} We use 15 epochs as the main setting and report 5, 10, and 20 epochs as a sensitivity analysis on the same data (Appendix~\ref{app:epoch}). The learning rate ($2 \times 10^{-5}$) and batch size (4) were set in early exploratory runs and were not changed after the reported results were seen. No early stopping or checkpoint selection was used, so predictions come from the final epoch. Because there is no separate development set, the main accuracies may be slightly optimistic.

\smallskip
\noindent\textbf{Conditions.} We compare four input conditions:
\begin{itemize}[itemsep=-2pt, topsep=4pt]
    \item \emph{Native:} the premise and hypothesis in the variety itself.
    \item \emph{English gloss:} the same model and training setup applied to the English glosses.
    \item \emph{Hypothesis-only:} the model sees only the hypothesis and not the premise.
    \item \emph{Premise-only:} the model sees only the premise and not the hypothesis.
\end{itemize}

\noindent The glosses are shared by the two varieties and are translations of the original material, not an independently authored English NLI corpus. The English gloss condition is therefore a single approximate reference condition, not a matched English benchmark for comparison.

The hypothesis-only and premise-only conditions are diagnostics: they test whether the label can be predicted from only one of the two texts. Hypothesis-only baselines are a standard check for annotation artifacts in NLI data \citep{gururangan2018,poliak2018}. The hypothesis-only model is also run in the transfer setting; premise-only transfer was not run.

\medskip
\noindent\textbf{Cross-variety transfer.} Transfer is evaluated on disjoint items. For each fold $k$, the model is trained on source-variety items outside fold \mbox{$k$ (91--92 items)} and tested on target-variety items in fold $k$ (22--23 items). The two subsets are scenario-matched, so the counterpart of a test item expresses the same scenario and label. Because the folds are shared by item number, the counterparts of the test items fall in fold $k$ and are excluded from training, so every test scenario is unseen. Within-variety and transfer results therefore use the same test items and the same number of training items, and differ only in the variety of the training data.

\medskip
\noindent\textbf{Counterpart-leak audit.} We additionally train on all 114 source-variety items and test on all 114 target-variety items. Here the counterpart of every test item is seen in training, so this setting is not a measure of transfer. We report it to illustrate how much leakage can inflate accuracy. It also trains on more items (114 instead of 91--92), so the gain over disjoint-item transfer cannot be attributed to leakage alone.

\subsection{NLI Evaluation Protocol}
\label{subsec:eval_protocol}

\noindent\textbf{What we report.} For each experiment and seed, we report two accuracies:
\begin{itemize}[itemsep=0pt, topsep=4pt]
    \item the \emph{mean five-fold accuracy}, with its standard deviation across folds;
    \item the \emph{pooled accuracy} over the 114 predictions made across the five folds, with the correct/total count.
\end{itemize}
The pooled accuracy is the value used in the main tables (where $\pm$ is the standard deviation over seeds) and in the statistical tests.

\medskip
\noindent\textbf{Accuracy.} Accuracy is the proportion of correct predictions among the evaluated examples. For seed $s$, the mean five-fold accuracy is
\begin{equation*}
    A_s = \frac{1}{5} \sum_{k=1}^{5} A_{s,k},
\end{equation*}

where $A_{s,k}$ is the accuracy on fold $k$. The pooled accuracy is
\begin{equation*}
    P_s = \frac{1}{114} \sum_{k=1}^{5} C_{s,k},
\end{equation*}
where $C_{s,k}$ is the number of correct predictions on fold $k$. The two differ slightly because the folds hold 23, 23, 23, 23, and 22 items. Chance level for three balanced classes is 33.3\%.

\medskip
\noindent\textbf{Seeds.} The three seeds are repeated runs on the same 114 examples, so they measure variation from initialization and training order, not independent \mbox{evidence}. We summarize them descriptively with the sample mean and standard deviation across them; the population standard deviation is given in brackets in the appendix tables. Summary statistics are computed from the underlying fold-level and item-level outputs before rounding, so recomputing from the rounded appendix values may not reproduce the displayed summaries exactly.

\medskip\noindent\textbf{Above-chance test.} We apply a one-sided exact binomial test of each seed's pooled correct count against a success probability of $1/3$, and report how many of the three seeds reach $p < 0.05$. A score of 47/114 is the smallest that passes (Appendix~\ref{app:stats}).

\medskip
\noindent\textbf{Paired comparison.} The question is whether the premise changes accuracy relative to the hypothesis alone. We compare the full-input and hypothesis-only models on the same items with an exact two-sided McNemar test, separately for each seed. This covers the two within-variety settings and the two transfer directions, for 12 comparisons in total \mbox{(Appendix~\ref{app:stats})}.

\medskip\noindent\textbf{How to read the results.} At this sample size, a single accuracy has about a 95\% interval of roughly $\pm 9$ points, so differences of a few points between conditions are within noise. ``Not significantly different'' should not be read as ``equivalent.'' Because the seeds share the same items, the count of seeds above chance describes consistency across runs, not independent replication. No correction for multiple comparisons is applied, so the $p$-values should be read as descriptive.

\subsection{NLI Results}
\label{subsec:results}

Table~\ref{tab:nli_results} summarizes the main NLI results in a single train/test matrix. The diagonal entries are within-variety five-fold cross-validation, and the off-diagonal entries are disjoint-item cross-variety transfer. Table~\ref{tab:nli_diagnostics} adds the hypothesis-only and premise-only diagnostics. The text below reports the within-variety results (including the English-gloss reference), cross-variety transfer, the diagnostics, and the counterpart-leak audit, in that order.

\begin{table}[H]
\caption{NLI baseline and cross-variety transfer results}
    \label{tab:nli_results}
    \centering
    \small
    \setlength{\tabcolsep}{4pt}
    \begin{adjustbox}{max width=\columnwidth}
    \begin{tabular}{lcc}
    \toprule
    \textbf{Train data} & \textbf{Test: Cameroon Pidgin} & \textbf{Test: Camfranglais} \\
    \midrule
    Cameroon Pidgin only & 44.7\% $\pm$ 2.3\% & 40.1\% $\pm$ 3.3\% \\
    Camfranglais only & 45.6\% $\pm$ 3.5\% & 46.2\% $\pm$ 1.8\% \\
    English glosses (reference) & 61.1\% $\pm$ 4.3\% & 61.1\% $\pm$ 4.3\% \\
    Chance level (3-class) & 33.3\% & 33.3\% \\
    \bottomrule
    \end{tabular}
    \end{adjustbox}
    
    \smallskip
    {\footnotesize\noindent\leftskip=0pt\rightskip=0pt\parfillskip=0pt plus 1fil
    \textit{Note.} Diagonal entries: pooled five-fold accuracy (114 predictions per seed). Off-diagonal entries: disjoint-item transfer, so no test item or its counterpart is seen in training. Mean $\pm$ sample SD over seeds 42, 123, and 2024. The English-gloss row is a shared reference, not a cross-variety experiment.\par}
   
\end{table}

\begin{table}[H]
    \caption{Accuracy (\%) with full input compared with hypothesis-only and premise-only input}
    \label{tab:nli_diagnostics}
    \centering
    \small
    \setlength{\tabcolsep}{4pt}
    \begin{adjustbox}{max width=\columnwidth}
    \begin{tabular}{lcccc}
    \toprule
    \textbf{Setting} & \shortstack{\textbf{Full}\\\textbf{input}} & \shortstack{\textbf{Hypothesis-}\\\textbf{only}} & \shortstack{\textbf{Premise-}\\\textbf{only}} & \shortstack{\textbf{McNemar}\\\textbf{significant}} \\
    \midrule
    Pidgin, within & 44.7 $\pm$ 2.3 & 41.8 $\pm$ 1.8 & 37.1 $\pm$ 1.8 & 0 of 3 \\
    Camfranglais, within & 46.2 $\pm$ 1.8 & 47.1 $\pm$ 5.0 & 38.3 $\pm$ 5.8 & 0 of 3 \\
    Pidgin $\rightarrow$ Camfranglais & 40.1 $\pm$ 3.3 & 44.2 $\pm$ 5.6 & -- & 0 of 3 \\
    Camfranglais $\rightarrow$ Pidgin & 45.6 $\pm$ 3.5 & 48.8 $\pm$ 2.7 & -- & 0 of 3 \\
    \bottomrule
    \end{tabular}
    \end{adjustbox}

\smallskip
   {\footnotesize\noindent\leftskip=0pt\rightskip=0pt\parfillskip=0pt plus 1fil
\textit{Note.} Pooled accuracy, mean $\pm$ sample SD over three seeds. ``McNemar significant.'' is the number of seeds in which full and hypothesis-only input differ (exact McNemar test, $p < 0.05$). Premise-only transfer was not run. Per-seed values and tests are in Appendices~\ref{app:nli_robustness}, \ref{app:transfer} and~\ref{app:stats}.\par}
\end{table}

\noindent\textbf{Within-variety results.} Cameroon Pidgin English obtained a descriptive mean accuracy of \mbox{44.7\% $\pm$ 2.3\%} and Camfranglais 46.2\% $\pm$ 1.8\%. The 1.5-point gap between the varieties is within seed variation, so we do not rank them by difficulty.

\noindent The English glosses obtained 61.1\% $\pm$ 4.3\%. All three (Pidgin, Camfranglais, gloss) are above the 33.3\% chance level in every seed (one-sided exact binomial test, $p < 0.05$ in 3 of 3 seeds), which indicates a learnable signal under the evaluated setup (see the diagnostics below).

Accuracy rises most from 5 to 10 epochs, and from 15 to 20 epochs it changes by $+1.2$ points (Pidgin) and $+2.3$ points (Camfranglais), which is within the seed-to-seed variation (Appendix~\ref{app:epoch}).

\medskip
\noindent\textbf{English gloss reference.} The English reference is about 15 to 16 points higher than the within-variety means. This comparison is treated with caution: the glosses are translations of the original material, so the result does not establish a language-specific representation gap in XLM-RoBERTa.

\medskip
\noindent\textbf{Cross-variety transfer.} A model trained on Pidgin reached 40.1\% $\pm$ 3.3\% on Camfranglais (above chance in 1 of 3 seeds), and a model trained on Camfranglais reached 45.6\% $\pm$ 3.5\% on Pidgin (above chance in 3 of 3 seeds). We therefore read transfer into Pidgin as above chance, though not above hypothesis-only transfer, and transfer into Camfranglais as suggestive only (one-sided binomial tests per seed; see Section~\ref{subsec:eval_protocol}).

\noindent These runs share test items and training-set size with the within-variety runs, so the two can be compared directly:
\begin{itemize}[itemsep=0pt, topsep=6pt]
    \item On Pidgin items, the Camfranglais-trained model (45.6\%) is about as accurate as a model trained on Pidgin itself (44.7\%).
    \item On Camfranglais items, the Pidgin-trained model is 6.1 points below a model trained on Camfranglais itself (40.1\% against 46.2\%).
\end{itemize}
With three seeds and 114 items, we do not claim that transfer across varieties is asymmetric.

\medskip
\noindent\textbf{Hypothesis-only and premise-only diagnostics.} A hypothesis-only model reaches 41.8\% (Pidgin) and 47.1\% (Camfranglais) within variety, close to the full-input models (44.7\% and 46.2\%). Paired exact McNemar tests find no significant difference between full and hypothesis-only input in any of the 12 seed-level comparisons (smallest $p = 0.212$). The full-input model is higher in 6 comparisons and lower in 6, and the mean differences range from $+2.9$ to $-4.1$ points.

The transfer results show the same pattern: hypothesis-only transfer (44.2\% and 48.8\%; \mbox{Table~\ref{tab:nli_diagnostics})} is at least as accurate as full-input transfer (40.1\% and 45.6\%). Premise-only models stay near chance (37.1\% and 38.3\%; above chance in 0 and 1 of 3 seeds), so the premise alone carries little of the label.
One possible cue is negation: no neutral hypothesis in either variety contains a negation word, whereas 16--29\% of the entailment and contradiction hypotheses do (Appendix~\ref{app:dataset_checks}). This does not by itself explain the hypothesis-only accuracy.

\medskip
\noindent\textbf{Counterpart-leak audit.} When the model is trained on all 114 source items and tested on all 114 target items, accuracy is 63.7\% $\pm$ 5.3\% (Pidgin $\rightarrow$ Camfranglais) and 60.5\% $\pm$ 8.0\% \mbox{(Camfranglais $\rightarrow$ Pidgin)}. These are about 24 and 15 points above disjoint-item transfer in the same directions. The gain is positive in all six seed pairs, between 5 and 31 additional correct predictions per seed.
 Because the counterpart of every test item has been seen in training, these values reflect leakage and a larger training set (114 instead of 91--92 items) as well as any transfer, and should not be read as evidence of cross-variety transfer.

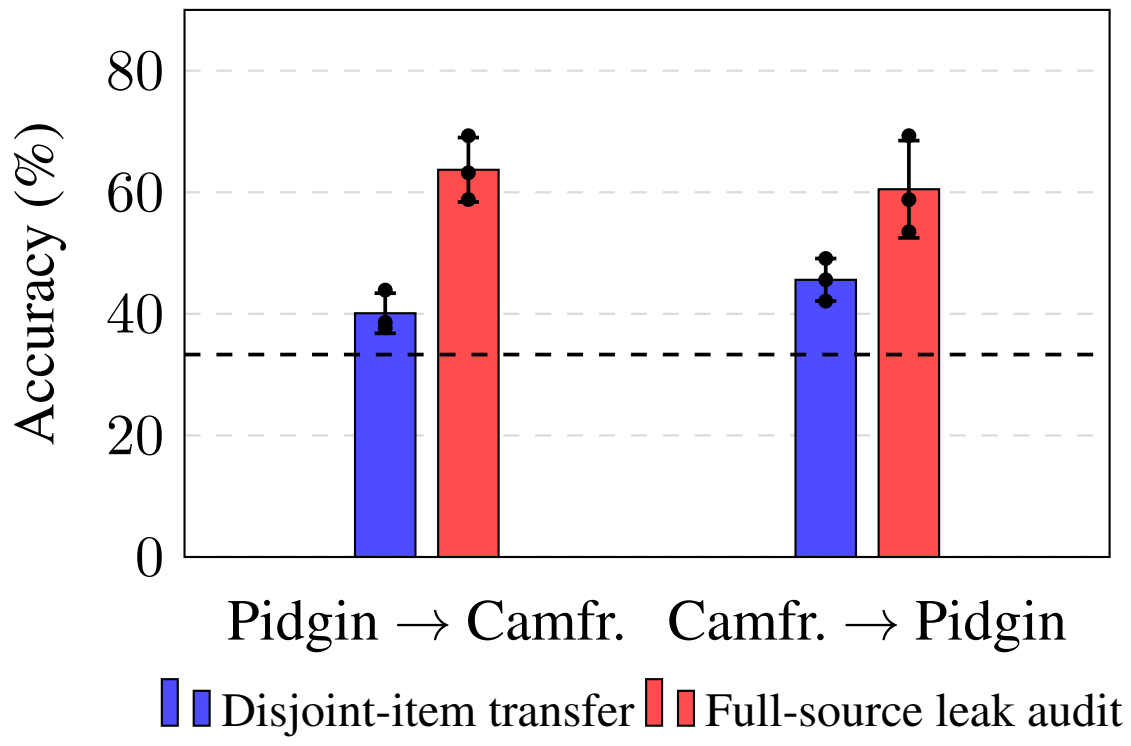
\begin{figure}[H]
\centering
\pgfplotsset{
  eb/.style={error bars/.cd, y dir=both, y explicit, error bar style={black, line width=0.7pt}, error mark options={black, rotate=90, mark size=2pt, line width=0.7pt}},
  seeddots/.style={only marks, mark=*, mark size=1.2pt, black, forget plot},
}
\begin{tikzpicture}
\begin{axis}[
    ybar,
    bar width=0.4cm,
    ylabel={Accuracy (\%)},
    xmin=-0.55, xmax=1.55,
    xtick={0,1}, xticklabels={{Pidgin $\rightarrow$ Camfr.},{Camfr.\ $\rightarrow$ Pidgin}},
    ymin=0, ymax=90, ytick={0,20,40,60,80},
    ymajorgrids=true,
    grid style={dashed,gray!30},
    legend style={at={(0.5,-0.2)}, anchor=north, legend columns=-1, draw=none, font=\footnotesize},
    axis line style={-},
    major tick style={draw=none},
    width=\columnwidth,
    height=5.2cm
]
\addplot+[fill=blue!70, draw=black, bar shift=-0.275cm, forget plot, eb] coordinates {(0,40.1) +- (0,3.3) (1,45.6) +- (0,3.5)};
\addplot+[fill=red!70, draw=black, bar shift=0.275cm, forget plot, eb] coordinates {(0,63.7) +- (0,5.3) (1,60.5) +- (0,8.0)};

\addlegendimage{ybar legend, fill=blue!70, draw=black}\addlegendentry{Disjoint-item transfer}
\addlegendimage{ybar legend, fill=red!70, draw=black}\addlegendentry{Full-source leak audit}

\addplot[seeddots, xshift=-0.275cm] coordinates {(0,37.7) (0,43.9) (0,38.6) (1,42.1) (1,45.6) (1,49.1)};
\addplot[seeddots, xshift=0.275cm] coordinates {(0,63.2) (0,69.3) (0,58.8) (1,69.3) (1,58.8) (1,53.5)};
\draw[dashed, line width=0.7pt] (axis cs:-0.55,33.3) -- (axis cs:1.55,33.3);
\end{axis}
\end{tikzpicture}
\caption{Disjoint-item transfer versus the counterpart-leak audit, which trains on all 114 source items (accuracy, \%). Bars show the mean over seeds 42, 123, and 2024, whiskers the sample SD, and dots the individual seeds. The dashed line is the 33.3\% chance level.}
\label{fig:transfer}
\end{figure}

\subsection{Speech Recognition Robustness Check}
\label{sec:asr}
The premises in this resource are spoken, so we also evaluated how well a general-purpose multilingual automatic speech recognition system transcribes the recordings, to see whether current tools can already handle these varieties (speech-check path in Figure~\ref{fig:pipeline}). We used Whisper \citep{radford2022} without task-specific fine-tuning, in three sizes (Base, Small, and Large-v3, which is a later release of the same model family) and three language conditions: automatic language identification, language forced to English, and language forced to French. Each condition was run on all 114 recordings per variety (2,052 transcripts in total), and Word Error Rate (WER) was measured against the reference premise texts.  This experiment serves as an audio-grounded robustness check rather than a dedicated ASR benchmark.

In general, Word Error Rate is defined as:
\begin{equation}
    \mathrm{WER} = \frac{S + D + I}{N}
\end{equation}
where $S$, $D$, and $I$ denote the number of substitutions, deletions, and insertions, and $N$ is the total number of words in the reference transcription. WER can exceed 100\% when the accumulated \mbox{errors} exceed the reference word count.

\medskip
\noindent\textbf{Setup.} For each variety $v$, the speech check uses the 114 audio recordings $a_i$ and their reference premise texts $r_i$:
\[
  \mathcal{D}_v = \{(a_i,\, r_i)\}_{i=1}^{114}
\]
For model $m$ (Base, Small, or Large-v3), let $\hat{r}_i^{(m)}$ be its raw transcription of $a_i$. We call this a \emph{transcription}, not a hypothesis, to avoid confusion with the NLI hypothesis.

\smallskip
\noindent\textbf{Text normalization.} Let $\phi(\cdot)$ be the normalization function, applied identically to the reference and the transcription (Appendix~\ref{app:asr}):

\begin{itemize}[itemsep=-2pt, topsep=6pt]
    \item Whisper language tags are removed.
    \item Text is converted to Unicode NFC and lowercased.
    \item Apostrophes are kept.
    \item Every other character that is not a letter, digit, or combining mark becomes a space.
\end{itemize}

\noindent\textbf{Word error rate.} Adapted to our data, a minimum-edit-distance alignment between $\phi(r_i)$ and $\phi(\hat{r}_i^{(m)})$ gives the substitutions $S_i$, deletions $D_i$, and insertions $I_i$ for each recording. With $N_i$ the number of words in $\phi(r_i)$, the corpus-level WER pools errors over the whole subset:
\begin{equation}
    \mathrm{WER}_v^{(m)} = \frac{\sum_{i=1}^{114} (S_i + D_i + I_i)}{\sum_{i=1}^{114} N_i}
\end{equation}
The reference denominator is 1,055 words for Cameroon Pidgin and 1,096 for Camfranglais. WER ranges are 95\% bootstrap intervals over the 114 recordings.

\smallskip
\noindent\textbf{Additional metrics.} Because neither variety has a standard spelling, WER penalizes a correct recognition that is spelled differently from the reference, so CER is less sensitive to spelling variation. We report CER on the same normalized text, together with MER, WIL, and WIP (definitions in Appendix~\ref{app:asr}).

\smallskip
\noindent\textbf{Language-identification counts.} Under automatic decoding, let $\ell_i^{(m)}$ be the language tag that model $m$ predicts for recording $a_i$. The number of distinct tags counts how many different languages the model assigns:
\[
  T_v^{(m)} = \bigl|\{\ell_i^{(m)} : i = 1, \dots, 114\}\bigr|
\]

\begin{table}[H]
    \caption{Zero-shot ASR results (normalized WER \%, with CER \% in brackets)}
    \label{tab:asr_results}
    \centering
    \small
    \setlength{\tabcolsep}{4pt}
    \begin{adjustbox}{max width=\columnwidth}
    \begin{tabular}{llccc}
    \toprule
    \textbf{Variety} & \textbf{Condition} & \shortstack{\textbf{Whisper}\\\textbf{Base}} & \shortstack{\textbf{Whisper}\\\textbf{Small}} & \shortstack{\textbf{Whisper}\\\textbf{Large-v3}} \\
    \midrule
    Cameroon Pidgin & Automatic & 91.8 (61.9) & 95.7 (70.2) & 80.6 (45.6) \\
    & Forced English & 80.3 (48.1) & 74.1 (39.0) & 66.6 (32.3) \\
    & Forced French & 136.2 (116.3) & 98.3 (72.6) & 98.3 (68.0) \\
    \midrule
    Camfranglais & Automatic & 90.1 (64.5) & 77.1 (52.5) & 64.9 (33.4) \\
    & Forced English & 100.2 (76.0) & 98.2 (68.4) & 93.9 (62.4) \\
    & Forced French & 79.3 (54.8) & 61.8 (39.3) & 45.5 (26.3) \\
    \bottomrule
    \end{tabular}
    \end{adjustbox}

    \medskip
    {\footnotesize\noindent\leftskip=0pt\rightskip=0pt\parfillskip=0pt plus 1fil
    \textit{Note.} Matched zero-shot conditions using all 114 recordings per variety. Values are corpus-level WER with CER in brackets, computed on normalized text. Full metrics and WER ranges are in Appendix~\ref{app:asr}.\par}
\end{table}

\label{fig:transfer}

\noindent\textbf{Findings.} Large-v3 helps, but does not solve the problem: it has the lowest WER in every row of Table~\ref{tab:asr_results} (tied with Small for Pidgin with forced French). The best results are:
\begin{itemize}[itemsep=-2pt, topsep=4pt]
    \item Camfranglais, forced French: 45.5\% WER (26.3\% CER);
    \item Pidgin, forced English: 66.6\% WER (32.3\% CER).
\end{itemize}
The ranges of the Pidgin automatic results for Base and Small overlap, so those two models cannot be ranked, and the Pidgin forced-French ranges for Small and Large-v3 include 100\%.

\medskip
\noindent\textbf{Language identification.} Under automatic decoding, Whisper rarely identifies Pidgin as English. Large-v3 tags 53 of 114 Pidgin recordings as Yoruba and 40 as Swahili, and only 9 as English. Camfranglais is most often tagged as French (54 of 114), but 24 recordings are tagged as Haitian Creole (Appendix~\ref{app:asr_lid}). Across all recordings, the number of distinct tags is 20 (Base), 11 (Small), and 9 (Large-v3) for Pidgin, and 23 (Base), 17 (Small), and 17 (Large-v3) for Camfranglais.

\smallskip
\noindent\textbf{Script errors.} Many transcripts also come out in the wrong script. For Pidgin under automatic decoding, 26 (Base), 34 (Small), and 14 (Large-v3) of the 114 transcripts contain non-Latin letters. Forcing the matching language largely removes this in both varieties (Appendix~\ref{app:asr_script}). 

\smallskip
\noindent\textbf{Forced versus automatic language.} Forcing the matching language lowers WER for every model and in both varieties, for example from 64.9\% to 45.5\% for Camfranglais with Large-v3. The best-per-variety results assume that the language of the recording is known in advance. Without that knowledge, the automatic setting applies, giving 80.6\% WER for Pidgin and 64.9\% for Camfranglais with Large-v3, the best model.

\medskip
\noindent\textbf{Error types.} The transcripts show several kinds of error: language misidentification, output in a different script, wrong substitutions of local vocabulary, and mismatches with the non-standard reference spelling (Appendix~\ref{app:asr_examples}).

\subsection{Interpretation}
\label{subsec:interpretation}

\begingroup
The NLI and ASR experiments examine two \mbox{separate} paths. The NLI experiments ask whether semantic inference can be learned from the available textual representations. The ASR experiments ask how hard it is for a general-purpose speech recognizer to recover the premise text from raw speech recordings.

The within-variety NLI results are above the 33.3\% chance level (44.7\% and 46.2\%), but they do not show that the model uses the relation between premise and hypothesis. A hypothesis-only model scores about the same (41.8\% and 47.1\%), and paired tests do not separate the two. The most cautious reading is that, here, hypothesis wording may account for much of the above-chance \mbox{accuracy}. This is a property of the current items, and it points to what a larger release should control, such as having independent annotators label the hypotheses without seeing the intended label.

The cross-variety results show the same pattern. Under disjoint-item splits, transfer into Camfranglais (40.1\%) is 6.1 points below within-variety Camfranglais accuracy (46.2\%), transfer into Pidgin (45.6\%) is close to within-variety Pidgin accuracy (44.7\%), and neither exceeds hypothesis-only transfer (44.2\% and 48.8\%). This pattern therefore does not demonstrate that inference ability transfers between the varieties. The higher full-source numbers (63.7\% and 60.5\%) were obtained with each test item's counterpart in training, which is consistent with counterpart overlap, although the audit also trains on more items (114 instead of 91--92). None of this bears on the linguistic relation between Cameroon Pidgin and Camfranglais.

The ASR results give a complementary view of the spoken path. With automatic language identification, Large-v3 reaches 80.6\% normalized WER for Pidgin and 64.9\% for Camfranglais. Forcing the matching language helps, but not enough (best 66.6\% and 45.5\%). For Pidgin, language identification rarely returns English (9 of 114 recordings with Large-v3). For Camfranglais, the largest gain comes from forcing French with the largest model (64.9\% to 45.5\%), which suggests that language misidentification accounts for a substantial part of the error. The remaining error may reflect code-switching, local vocabulary, and non-standard spelling, which these experiments do not separate (Appendix~\ref{app:asr_examples}). Together with the frequent language misidentification, these results indicate that recovering a reliable textual representation from speech is difficult for the evaluated off-the-shelf multilingual recognizers.

The ASR results do not explain the NLI results: the NLI baseline uses reference text rather than Whisper-generated transcripts, so the present experiments do not measure how transcription errors propagate into semantic inference. The defensible joint conclusion is that the dataset exposes two complementary challenges: semantic inference from the textual representation, including how far the labels depend on hypothesis wording, and reliable conversion of spoken input into that representation.

This separation also clarifies the role of this resource. The current release (v1.0) establishes an initial dataset, annotation procedure, and baseline evaluation framework, not an end-to-end spoken-NLI solution.
\endgroup

\medskip
\section{Future Work}
\label{sec:future_work}
This pilot is a stepping stone. A future version (v2.0) is envisioned as a larger, more diverse resource in which the two varieties no longer share the same scenarios. The planned extensions fall into five parts:

\medskip
\noindent\textbf{Speakers, Annotation and Splits.} We plan to record multiple speakers across demographic groups and Cameroonian regions, using varied devices and acoustic conditions, to test robustness to accent, dialect, and recording-channel variation, and to use independent external annotators so that inter-annotator agreement (e.g., Cohen's $\kappa$ or Fleiss' $\kappa$) can be reported under documented consent procedures. Version 2.0 also aims to provide held-out development and test partitions.

\medskip
\noindent\textbf{Unmatched Contexts and Cue Mitigation.} To prevent models from exploiting parallel alignments, v2.0 aims to introduce distinct, unmatched scenarios for Camfranglais that decouple it from the Pidgin subset. Hypotheses will be written by several annotators and then labeled by independent annotators who do not see the intended label, and the use of negation words will be balanced across labels to reduce phrase-level artifacts.

\medskip
\noindent\textbf{ASR Error Propagation and Model Scope.} We plan to feed predicted ASR transcripts into the NLI models to measure how word- and character-level errors affect inference. Because neither variety has a standardized script, this will require unified transcription conventions, which could also support fine-tuning speech models on these varieties.    We also plan to evaluate larger multilingual encoders and prompted language models.

\medskip
\noindent\textbf{End-to-End Spoken NLI.} We intend to record spoken hypotheses alongside spoken premises, \mbox{moving} from text-based NLI to inference directly from speech, without intermediate transcription.

\medskip
\noindent\textbf{Additional Varieties.} As the collection expands, we aim to add other oral-first Cameroonian language varieties to the evaluation suite, supporting research toward voice-first technologies for under-resourced communities.

\section{Conclusion}
\label{sec:conclusion}
We presented CamSpokenNLI, a pilot spoken Natural Language Inference resource for Cameroon Pidgin English and Camfranglais, with 114 balanced scenario-matched items per variety. Fine-tuned XLM-RoBERTa-base reaches 44.7\% on Pidgin and 46.2\% on Camfranglais under five-fold cross-validation, above the 33.3\% chance level, and disjoint-item transfer reaches 40.1\% from Pidgin to Camfranglais and 45.6\% in the reverse direction. Diagnostic checks show that these scores should be read with care: a hypothesis-only baseline performs about as well as full input, and letting counterpart items into training raises accuracy by about 15 to 24 points, illustrating how counterpart leakage can inflate transfer results.

Evaluated separately from NLI, zero-shot Whisper remains far from adequate on both varieties: its best normalized WER is 66.6\% on Pidgin and 45.5\% on Camfranglais, reached only when the matching language is forced.

These results are exploratory because the release is small, single-speaker, single-annotator and scenario-matched, with no held-out test partition. It is therefore a seed resource, not a benchmark. We release it with its code to support larger, multi-speaker, independently annotated successors and, eventually, end-to-end spoken NLI for these under-resourced varieties.

\section*{Limitations}
\label{sec:limitations}
As a pilot resource, CamSpokenNLI has several limitations that bound its scope:

\medskip
\noindent\textbf{Scale and Coverage.} Each variety has only 114 examples, so evaluation uses stratified five-fold cross-validation instead of an independent held-out test set. At this scale, a single seed needs 47 correct predictions (41.2\%) to exceed chance at $p < 0.05$, and repeating seeds on the same items does not give independent evidence. All items come from one person, the author, a native of Buea, who wrote the scenarios from everyday life in Buea/Molyko, recorded all the audio, and labeled every item. The dataset therefore does not capture regional, demographic, accent, or topical variation, and inter-annotator agreement cannot be computed.

\medskip
\noindent\textbf{Dataset Construction and Artifacts.} The two subsets share the same 114 scenarios, English glosses, and gold labels. They are not independent collections, so the Camfranglais results are not an independent replication of the Pidgin findings, and the glosses are not an independent control. Hypotheses were written together with their target labels, so style may leak the label; for example, no neutral hypothesis contains a negation word. This does not by itself explain the hypothesis-only accuracy, but the hypothesis-only baseline performs comparably to the full-input model and paired tests do not separate them. The accuracies should \mbox{therefore} not be read as proof of semantic inference, nor as measures of mutual intelligibility or structural similarity between the two varieties.

\medskip
\noindent\textbf{Modeling Scope and Training.} We fine-tuned only \mbox{XLM-RoBERTa-base}; larger multilingual encoders, prompted language models, and instruction-tuned models were not evaluated. The learning rate and batch size were set in early exploratory runs, and there is no separate development set, which may introduce a mild optimistic bias. We kept \mbox{15 epochs} as the main setting; a 20-epoch run scored higher by $+1.2$ points for Pidgin and $+2.3$ points for Camfranglais, within seed variation. Training beyond 20 epochs was not explored.

\medskip
\noindent\textbf{Audio and Speech Recognition.} The NLI baselines use the reference text and do not use the audio; the recordings are evaluated only through a separate Whisper check. Only the premises are recorded; the hypotheses exist as text only. The recordings were made on a consumer mobile phone, not across varied acoustic conditions. The Whisper results (Base, Small, and Large-v3) are an \mbox{out-of-the-box} zero-shot diagnostic, not a fine-tuned ASR benchmark, and only English and French decoding were forced. The best word error rates remain high: 66.6\% for Pidgin (forced English) and 45.5\% for Camfranglais (forced French), both with Large-v3. Because neither variety has a standardized orthography, these figures partly reflect spelling mismatch rather than incorrect word recognition. Since the NLI experiments use reference text, the effect of ASR errors on a full spoken-NLI pipeline remains unmeasured in this study.

\section*{Broader Impact and Ethics Statement}
\phantomsection\label{sec:ethics}
\noindent\textbf{Broader Impact.} This resource is intended to support research on spoken language technology for Cameroon Pidgin English and Camfranglais, two under-resourced varieties. It may inform future work on local conversational agents, spoken information retrieval, and voice-driven services. However, because this release is a single-speaker, single-annotator pilot, it is unsuited to benchmarking production systems, and should not be used to draw general sociolinguistic conclusions about the speaker communities. Systems built on it may work worse for other speakers and regions, and should not be deployed without evaluation on the intended users.

\medskip
\noindent\textbf{Consent, Licensing, and Responsible Data Use.} All sentences, recordings, and annotations were created by the author, who consents to their release under CC BY-NC 4.0 for non-commercial research.
The scenarios come from everyday life, including utilities, local transport, marketplaces, basic education, and financial transactions. As far as the author has checked, the data contain no sensitive or offensive content and no personally identifiable information. Any future multi-speaker collection will require documented consent and institutional ethical review.

\medskip
\noindent\textbf{Conflict of Interest.} The dataset was developed alongside \mbox{LocalHands Africa}, a voice-first service-exchange platform created by the author, whose intended domain overlaps with the tasks \mbox{studied} here. No data in the v1.0 release come from \mbox{LocalHands Africa}, and the experiments use only general-purpose models. To support independent verification, the dataset, evaluation results, and scripts are made available for research use.

\section*{\textbf{\NoCaseChange{Data and Code Availability}}}
\phantomsection\label{sec:data}
The CamSpokenNLI dataset, including its audio recordings and linguistic annotations, was developed under \textbf{\mbox{LAB~237}} and is available at \href{https://huggingface.co/datasets/lab-237/camspoken}{\nolinkurl{huggingface.co/datasets/lab-237/camspoken}} for non-commercial use under the Creative Commons Attribution-NonCommercial 4.0 International License (CC BY-NC 4.0). The data processing pipeline, reproducible notebooks, and experimental results are available at \href{https://github.com/lab-237/camspoken}{\nolinkurl{github.com/lab-237/camspoken}} under the PolyForm Noncommercial License 1.0.0. Both repositories are tagged \texttt{v1.0}.

\section*{\textbf{\NoCaseChange{Acknowledgments}}}

The author acknowledges the open-source model and software communities upon whose work this study builds, and Google Colab for providing the computational resources.


\clubpenalty=150
\widowpenalty=150
\renewcommand{\bibfont}{\fontsize{10.5}{12.5}\selectfont}
\setlength{\bibsep}{3pt plus 1pt minus 1pt}
\sloppy
\bibliography{references}

@inproceedings{adelani2025,
  author    = {David Ifeoluwa Adelani and Jessica Ojo and Israel Abebe Azime and Jian Yun Zhuang and Jesujoba O. Alabi and Xuanli He and Millicent Ochieng and Sara Hooker and Andiswa Bukula and En-Shiun Annie Lee and others},
  year      = {2025},
  title     = {{IrokoBench}: A New Benchmark for {African} Languages in the Age of Large Language Models},
  booktitle = {Proceedings of the 2025 Conference of the Nations of the Americas Chapter of the Association for Computational Linguistics: Human Language Technologies (Volume 1: Long Papers)},
  pages     = {2732--2757},
  publisher = {Association for Computational Linguistics},
  url       = {https://aclanthology.org/2025.naacl-long.139/}
}

@inproceedings{armstrong2022,
  author    = {Ruth-Ann Armstrong and John Hewitt and Christopher Manning},
  year      = {2022},
  title     = {{JamPatoisNLI}: A {Jamaican} Patois Natural Language Inference Dataset},
  booktitle = {Findings of the Association for Computational Linguistics: EMNLP 2022},
  pages     = {5307--5320},
  address   = {Abu Dhabi, United Arab Emirates},
  publisher = {Association for Computational Linguistics},
  doi       = {10.18653/v1/2022.findings-emnlp.389},
  url       = {https://aclanthology.org/2022.findings-emnlp.389/}
}

@book{ayafor2017,
  author    = {Miriam Ayafor and Melanie Green},
  year      = {2017},
  title     = {{Cameroon Pidgin English}: A Comprehensive Grammar},
  series    = {London Oriental and African Language Library},
  volume    = {20},
  publisher = {John Benjamins Publishing Company},
  address   = {Amsterdam},
  doi       = {10.1075/loall.20}
}

@inproceedings{bowman2015,
  author    = {Samuel R. Bowman and Gabor Angeli and Christopher Potts and Christopher D. Manning},
  year      = {2015},
  title     = {A Large Annotated Corpus for Learning Natural Language Inference},
  booktitle = {Proceedings of the 2015 Conference on Empirical Methods in Natural Language Processing},
  pages     = {632--642},
  address   = {Lisbon, Portugal},
  publisher = {Association for Computational Linguistics},
  url       = {https://aclanthology.org/D15-1075/}
}

@inproceedings{conneau2018,
  author    = {Alexis Conneau and Ruty Rinott and Guillaume Lample and Adina Williams and Samuel R. Bowman and Holger Schwenk and Veselin Stoyanov},
  year      = {2018},
  title     = {{XNLI}: Evaluating Cross-lingual Sentence Representations},
  booktitle = {Proceedings of the 2018 Conference on Empirical Methods in Natural Language Processing},
  pages     = {2475--2485},
  address   = {Brussels, Belgium},
  publisher = {Association for Computational Linguistics},
  url       = {https://aclanthology.org/D18-1269/}
}

@inproceedings{conneau2020,
  author    = {Alexis Conneau and Kartikay Khandelwal and Naman Goyal and Vishrav Chaudhary and Guillaume Wenzek and Francisco Guzm{\'a}n and Edouard Grave and Myle Ott and Luke Zettlemoyer and Veselin Stoyanov},
  year      = {2020},
  title     = {Unsupervised Cross-lingual Representation Learning at Scale},
  booktitle = {Proceedings of the 58th Annual Meeting of the Association for Computational Linguistics},
  pages     = {8440--8451},
  address   = {Online},
  publisher = {Association for Computational Linguistics},
  doi       = {10.18653/v1/2020.acl-main.747}
}

@inproceedings{deshmukh2025,
  author    = {Deshmukh, Soham and Han, Shuo and Bukhari, Hazim and Elizalde, Benjamin and Gamper, Hannes and Singh, Rita and Raj, Bhiksha},
  title     = {Audio Entailment: Assessing Deductive Reasoning for Audio Understanding},
  booktitle = {Proceedings of the AAAI Conference on Artificial Intelligence},
  volume    = {39},
  pages     = {23769--23777},
  year      = {2025},
  doi       = {10.1609/aaai.v39i22.34548},
  url       = {https://doi.org/10.1609/aaai.v39i22.34548}
}

@article{echu2003,
  author  = {George Echu},
  year    = {2003},
  title   = {Influence of {Cameroon Pidgin English} on the Linguistic and Cultural Development of the {French} Language},
  journal = {IULC Working Papers},
  volume  = {3},
  number  = {1},
  url     = {https://scholarworks.iu.edu/journals/index.php/iulcwp/article/view/25950}
}

@article{echu2006,
  author  = {George Echu},
  year    = {2006},
  title   = {Pidginization of the {French} Language in {Cameroon}},
  journal = {TRANS: Internet-Zeitschrift f{\"u}r Kulturwissenschaften},
  volume  = {16},
  url     = {https://www.inst.at/trans/16Nr/01_5/echu16.htm}
}

@misc{kondoro2026,
  author       = {Alfred Malengo Kondoro and Sharon Ibejih and Cynthia Amol and others},
  year         = {2026},
  title        = {{Afri-XNLI+}: Extending the {XNLI} Dataset with New {African} Language Translations},
  howpublished = {Dataset, Tonative (Hugging Face)},
  doi          = {10.57967/hf/7718}
}

@article{kouega2003,
  author  = {Jean-Paul Kouega},
  year    = {2003},
  title   = {{Camfranglais}: A Novel Slang in {Cameroon} Schools},
  journal = {English Today},
  volume  = {19},
  number  = {2},
  pages   = {23--29},
  doi     = {10.1017/S0266078403002050}
}

@inproceedings{muhammad2022,
  author    = {Shamsuddeen Hassan Muhammad and David Ifeoluwa Adelani and Sebastian Ruder and Ibrahim Sa'id Ahmad and Idris Abdulmumin and Bello Shehu Bello and Monojit Choudhury and Chris Chinenye Emezue and Saheed Salahudeen Abdullahi and Anuoluwapo Aremu and Al{\' i}pio Jorge and Pavel Brazdil},
  year      = {2022},
  title     = {{NaijaSenti}: A {Nigerian} {Twitter} Sentiment Corpus for Multilingual Sentiment Analysis},
  booktitle = {Proceedings of the Thirteenth Language Resources and Evaluation Conference},
  pages     = {590--602},
  address   = {Marseille, France},
  publisher = {European Language Resources Association},
  url       = {https://aclanthology.org/2022.lrec-1.63/}
}

@inproceedings{okocha2026,
  author    = {Chibuzor Okocha and Christan Grant},
  year      = {2026},
  title     = {{Afrispeech} Semantics: Evaluating Audio-Semantic Reasoning in Spoken Language Models Across Domains and Accents},
  booktitle = {Findings of the Association for Computational Linguistics: ACL 2026},
  pages     = {6909--6928},
  address   = {San Diego, California, United States},
  publisher = {Association for Computational Linguistics},
  doi       = {10.18653/v1/2026.findings-acl.343},
  url       = {https://aclanthology.org/2026.findings-acl.343/}
}

@article{ozon2017spoken,
  title={The spoken corpus of {Cameroon Pidgin English}},
  author={Oz{\'o}n, Gabriel and Ayafor, Miriam and Green, Melanie and Fitzgerald, Sarah},
  journal={World Englishes},
  volume={36},
  number={3},
  pages={427--447},
  year={2017},
  publisher={Wiley Online Library},
  doi={10.1111/weng.12280}
}

@misc{radford2022,
  author       = {Alec Radford and Jong Wook Kim and Tao Xu and Greg Brockman and Christine McLeavey and Ilya Sutskever},
  year         = {2022},
  title        = {Robust Speech Recognition via Large-Scale Weak Supervision},
  howpublished = {arXiv preprint arXiv:2212.04356},
  url          = {https://arxiv.org/abs/2212.04356}
}

@incollection{schroder2013,
  author    = {Anne Schr{\"o}der},
  year      = {2013},
  title     = {{Cameroon Pidgin English}},
  booktitle = {The Survey of Pidgin and Creole Languages, Volume I: English-based and Dutch-based Languages},
  editor    = {Susanne Maria Michaelis and Philippe Maurer and Martin Haspelmath and Magnus Huber},
  pages     = {185--193},
  publisher = {Oxford University Press},
  address   = {Oxford},
  url       = {https://apics-online.info/surveys/18}
}

@article{ubanako2021,
  author  = {Valentine Njende Ubanako},
  year    = {2021},
  title   = {Is {Camfranglais} a New Language? A Review of Current Opinions},
  journal = {International Linguistics Research},
  volume  = {4},
  number  = {1},
  pages   = {36--42},
  doi     = {10.30560/ilr.v4n1p36}
}

@inproceedings{williams2018,
  author    = {Adina Williams and Nikita Nangia and Samuel R. Bowman},
  year      = {2018},
  title     = {A Broad-Coverage Challenge Corpus for Sentence Understanding through Inference},
  booktitle = {Proceedings of the 2018 Conference of the North American Chapter of the Association for Computational Linguistics: Human Language Technologies, Volume 1 (Long Papers)},
  pages     = {1112--1122},
  address   = {New Orleans, Louisiana},
  publisher = {Association for Computational Linguistics},
  url       = {https://aclanthology.org/N18-1101/}
}

@article{yakpo2024,
  author  = {Kofi Yakpo},
  year    = {2024},
  title   = {{West African Pidgin}: World Language Against the Grain},
  journal = {Africa Spectrum},
  volume  = {59},
  number  = {2},
  pages   = {180--203},
  doi     = {10.1177/00020397241263364}
}

@incollection{faraclas2013,
  author    = {Nicholas G. Faraclas},
  year      = {2013},
  title     = {{Nigerian Pidgin}},
  booktitle = {The Survey of Pidgin and Creole Languages, Volume I: English-based and Dutch-based Languages},
  editor    = {Susanne Maria Michaelis and Philippe Maurer and Martin Haspelmath and Magnus Huber},
  pages     = {176--184},
  publisher = {Oxford University Press},
  address   = {Oxford},
  url       = {https://apics-online.info/surveys/17}
}

@inproceedings{gururangan2018,
  author    = {Suchin Gururangan and Swabha Swayamdipta and Omer Levy and Roy Schwartz and Samuel R. Bowman and Noah A. Smith},
  year      = {2018},
  title     = {Annotation Artifacts in Natural Language Inference Data},
  booktitle = {Proceedings of the 2018 Conference of the North American Chapter of the Association for Computational Linguistics: Human Language Technologies, Volume 2 (Short Papers)},
  pages     = {107--112},
  address   = {New Orleans, Louisiana},
  publisher = {Association for Computational Linguistics},
  url       = {https://aclanthology.org/N18-2017/}
}

@inproceedings{poliak2018,
  author    = {Adam Poliak and Jason Naradowsky and Aparajita Haldar and Rachel Rudinger and Benjamin Van Durme},
  year      = {2018},
  title     = {Hypothesis Only Baselines in Natural Language Inference},
  booktitle = {Proceedings of the Seventh Joint Conference on Lexical and Computational Semantics},
  pages     = {180--191},
  address   = {New Orleans, Louisiana},
  publisher = {Association for Computational Linguistics},
  url       = {https://aclanthology.org/S18-2023/}
}

@book{lewis2014,
  editor    = {M. Paul Lewis and Gary F. Simons and Charles D. Fennig},
  year      = {2014},
  title     = {{Ethnologue}: Languages of the World},
  edition   = {17th},
  publisher = {SIL International},
  address   = {Dallas, Texas},
  url       = {http://www.ethnologue.com}
}
\fussy

\clearpage
\onecolumn 
\appendix
\section*{Appendices}

\section{Experimental Configuration}
\label{app:config}
This appendix lists the settings used for all reported NLI and ASR runs.

\subsection{XLM-RoBERTa NLI Configuration}
\label{app:config_xlmr}
To facilitate reproducibility of the NLI experiments, we detail the fine-tuning configuration. All experiments were run on a single NVIDIA Tesla T4 GPU via Google Colab.

\begin{itemize}[itemsep=0pt, topsep=2pt]
    \item \textbf{Base model:} \texttt{xlm-roberta-base}
    \item \textbf{Learning rate:} $2 \times 10^{-5}$, linear schedule
    \item \textbf{Optimizer:} AdamW; weight decay $0.0$
    \item \textbf{Batch size:} 4 per-device train, 8 per-device eval
    \item \textbf{Precision:} mixed precision (fp16)
    \item \textbf{Epochs:} 15 (the main setting); 5, 10, and 20 epochs are reported as a sensitivity analysis (Appendix~\ref{app:epoch})
    \item \textbf{Maximum sequence length:} 64 tokens
    \item \textbf{Input (native condition):} the premise and hypothesis exactly as in the released CSV (no lowercasing or punctuation removal)
    \item \textbf{Seeds:} 42, 123, 2024 (three repeated runs per condition; descriptive mean and standard deviation of accuracy reported)
    \item \textbf{Folds:} one stratified five-fold assignment on item numbers 1--114, shared by both varieties (fold seed 42; Appendix~\ref{app:folds})
    \item \textbf{Cross-variety transfer:} disjoint-item. For each fold, the model trains on source-variety items outside that fold and is tested on target-variety items inside it, so the counterpart of a test item is never seen in training. Training on all 114 source items and testing on all 114 target items is reported only as a counterpart-leak audit (Appendix~\ref{app:tr_leak}).
\end{itemize}

\smallskip
\noindent\textit{Protocol note. The learning rate and batch size were set in early exploratory runs, without a separate development set, and were not changed after the reported results were seen. No early stopping or checkpoint selection was used, so predictions always come from the final epoch. The seeds are repeated runs on the same 114 items per variety rather than independent test sets, and the fold assignment is fixed for all seeds, conditions, and experiments. Hence, the seeds vary only the model initialization and training order. In the across-seed rows, $\pm$ is the sample standard deviation over the three seeds, with the population value in brackets (``pop.''); in the per-seed rows, it is the standard deviation across the five folds of that seed's cross-validation.}
\medskip

\subsection{Experimental Conditions and Run Counts}
\begin{center}
\small
\begin{tabular}{llc}
\toprule
\textbf{Experiment} & \textbf{Setting} & \textbf{Runs} \\
\midrule
Within-variety, native input & 2 varieties $\times$ 3 seeds $\times$ 5 folds & 30 \\
Within-variety, English gloss & 2 varieties $\times$ 3 seeds $\times$ 5 folds & 30 \\
Within-variety, hypothesis-only & 2 varieties $\times$ 3 seeds $\times$ 5 folds & 30 \\
Within-variety, premise-only & 2 varieties $\times$ 3 seeds $\times$ 5 folds & 30 \\
Epoch sensitivity, native input (5, 10, 20 epochs) & 2 varieties $\times$ 3 epoch settings $\times$ 3 seeds $\times$ 5 folds & 90 \\
Disjoint-item transfer (native and hyp.-only) & 2 directions $\times$ 2 inputs $\times$ 3 seeds $\times$ 5 folds & 60 \\
Counterpart-leak audit & 2 directions $\times$ 3 seeds & 6 \\
\midrule
\textbf{Total} & & \textbf{276} \\
\bottomrule
\end{tabular}
\end{center}
\medskip
\noindent\textit{The 15-epoch native runs serve as the 15-epoch point of the epoch study. The two English-gloss conditions use the same glosses, labels, and folds, so they give identical results; the two gloss conditions are counted separately in the table.}
\medskip

\subsection{Whisper ASR Configuration}
\begin{itemize}[itemsep=0pt, topsep=2pt]
    \item \textbf{Models:} OpenAI Whisper Base, Small, and Large-v3 (\texttt{openai-whisper} 20250625), zero-shot, with no fine-tuning
    \item \textbf{Conditions:} automatic language identification, forced English and forced French, for both varieties
    \item \textbf{Scale:} 18 blocks (2 varieties × 3 models × 3 language conditions) of 114 recordings, 2,052 transcripts
    \item \textbf{Decoding:} default settings, with the random seed set to 0 before every recording
    \item \textbf{Audio:} the 16 kHz mono recordings described in Appendix~\ref{app:dataset_checks}
\end{itemize}

\subsection{Software and Reproducibility}
Python 3.13.15, PyTorch 2.11.0 (CUDA 13.0), Transformers 5.16.1, Datasets 4.8.5, scikit-learn 1.6.1, SciPy 1.16.3, jiwer 4.0.0, NumPy 2.1.3, pandas 2.2.3, and FFmpeg (Colab build, version not recorded). The release includes SHA-256 fingerprints for the two dataset files, the fold assignment and every audio file, along with a run manifest. Every NLI run stores its 114 item-level predictions.
\smallskip

 \noindent Mean training loss over all optimization steps, for the 120 within-variety runs at 15 epochs, was 0.80 (Pidgin) and 0.67 (Camfranglais) on native text, 0.52 on the English gloss, 0.81 / 0.75 for hypothesis-only and 0.86 / 0.83 for premise-only input, against about 1.10 for uniform guessing; it was logged for these runs only. All NLI and ASR analyses were rerun identically from the saved files. The Whisper transcripts were generated once, and their exact reproducibility on a different GPU has not been tested.

\section{NLI Robustness and  Per-Seed Results}
\label{app:nli_robustness}
This appendix preserves the seed-level NLI evidence supporting the main-paper summaries. It functions as the audit trail for the small-sample experiments rather than as an alternative headline result table. $p$ is the one-sided exact binomial test of the pooled accuracy against chance (1/3). In the ``Across seeds'' row, the last column counts the seeds with $p < 0.05$.

\subsection{Cameroon Pidgin English -- Primary Configuration (15 epochs)}
\begin{center}
\small
\begin{tabular}{cccc}
\toprule
\textbf{Seed} & \textbf{Mean accuracy $\pm$ SD across folds} & \textbf{Pooled accuracy} & \textbf{$p$ (vs 1/3)} \\
\midrule
42 & 45.5\% $\pm$ 8.0\% & 52/114 (45.6\%) & 0.0043 \\
123 & 46.5\% $\pm$ 8.1\% & 53/114 (46.5\%) & 0.0024 \\
2024 & 42.1\% $\pm$ 4.8\% & 48/114 (42.1\%) & 0.0312 \\
\midrule
\textbf{Across seeds} & -- & \textbf{44.7\% $\pm$ 2.3\% (pop. 1.9\%)} & \textbf{3 of 3} \\
\bottomrule
\end{tabular}
\end{center}
\medskip
\noindent\textit{The pooled accuracy is computed over the 114 examples for each seed. The three seeds are repeated runs, not independent test sets.}
\medskip

\subsection{English Gloss Reference -- Cameroon Pidgin Material}
\begin{center}
\small
\begin{tabular}{cccc}
\toprule
\textbf{Seed} & \textbf{Mean accuracy $\pm$ SD across folds} & \textbf{Pooled accuracy} & \textbf{$p$ (vs 1/3)} \\
\midrule
42 & 64.0\% $\pm$ 7.8\% & 73/114 (64.0\%) & $< 0.0001$ \\
123 & 56.2\% $\pm$ 5.9\% & 64/114 (56.1\%) & $< 0.0001$ \\
2024 & 63.2\% $\pm$ 13.8\% & 72/114 (63.2\%) & $< 0.0001$ \\
\midrule
\textbf{Across seeds} & -- & \textbf{61.1\% $\pm$ 4.3\% (pop. 3.5\%)} & \textbf{3 of 3} \\
\bottomrule
\end{tabular}
\end{center}
\medskip
\noindent\textit{The English glosses are translations of the original Pidgin and Camfranglais material and therefore constitute an approximate reference condition rather than an independently authored English corpus. The glosses, labels and folds are the same for both varieties, so B.2 and B.4 give identical results by construction (Section~\ref{subsec:setup}).}
\medskip

\subsection{Camfranglais -- Primary Configuration (15 epochs)}

\vspace{0.8\baselineskip}
\begin{center}
\small
\begin{tabular}{cccc}
\toprule
\textbf{Seed} & \textbf{Mean accuracy $\pm$ SD across folds} & \textbf{Pooled accuracy} & \textbf{$p$ (vs 1/3)} \\
\midrule
42 & 45.5\% $\pm$ 10.0\% & 52/114 (45.6\%) & 0.0043 \\
123 & 44.7\% $\pm$ 7.3\% & 51/114 (44.7\%) & 0.0074 \\
2024 & 48.2\% $\pm$ 10.8\% & 55/114 (48.2\%) & 0.0007 \\
\midrule
\textbf{Across seeds} & -- & \textbf{46.2\% $\pm$ 1.8\% (pop. 1.5\%)} & \textbf{3 of 3} \\
\bottomrule
\end{tabular}
\end{center}
\medskip
\noindent\textit{The three seeds are repeated runs on the same 114 examples, not independent test sets.}
\medskip

\subsection{English Gloss Reference -- Camfranglais Material}

\vspace{0.8\baselineskip}
\begin{center}
\small
\begin{tabular}{cccc}
\toprule
\textbf{Seed} & \textbf{Mean accuracy $\pm$ SD across folds} & \textbf{Pooled accuracy} & \textbf{$p$ (vs 1/3)} \\
\midrule
42 & 64.0\% $\pm$ 7.8\% & 73/114 (64.0\%) & $< 0.0001$ \\
123 & 56.2\% $\pm$ 5.9\% & 64/114 (56.1\%) & $< 0.0001$ \\
2024 & 63.2\% $\pm$ 13.8\% & 72/114 (63.2\%) & $< 0.0001$ \\
\midrule
\textbf{Across seeds} & -- & \textbf{61.1\% $\pm$ 4.3\% (pop. 3.5\%)} & \textbf{3 of 3} \\
\bottomrule
\end{tabular}
\end{center}

\medskip

\noindent\textit{As in Appendix B.2, this condition is an approximate translated reference rather than an independently authored English corpus, and it gives the same results as B.2 by construction.}
\medskip

\subsection{Hypothesis-Only Baseline -- Cameroon Pidgin English}

\vspace{0.8\baselineskip}
\begin{center}
\small
\begin{tabular}{cccc}
\toprule
\textbf{Seed} & \textbf{Mean accuracy $\pm$ SD across folds} & \textbf{Pooled accuracy} & \textbf{$p$ (vs 1/3)} \\
\midrule
42 & 40.3\% $\pm$ 3.1\% & 46/114 (40.4\%) & 0.0696 \\
123 & 43.9\% $\pm$ 3.2\% & 50/114 (43.9\%) & 0.0123 \\
2024 & 41.2\% $\pm$ 4.9\% & 47/114 (41.2\%) & 0.0473 \\
\midrule
\textbf{Across seeds} & -- & \textbf{41.8\% $\pm$ 1.8\% (pop. 1.5\%)} & \textbf{2 of 3} \\
\bottomrule
\end{tabular}
\end{center}

\medskip

\noindent\textit{The model sees only the hypothesis. This shows how much of the native-input accuracy can be reached without the premise.}
\medskip

\subsection{Hypothesis-Only Baseline -- Camfranglais}
\vspace{0.8\baselineskip}
\begin{center}
\small
\begin{tabular}{cccc}
\toprule
\textbf{Seed} & \textbf{Mean accuracy $\pm$ SD across folds} & \textbf{Pooled accuracy} & \textbf{$p$ (vs 1/3)} \\
\midrule
42 & 52.5\% $\pm$ 15.5\% & 60/114 (52.6\%) & $< 0.0001$ \\
123 & 42.8\% $\pm$ 15.8\% & 49/114 (43.0\%) & 0.0199 \\
2024 & 45.5\% $\pm$ 8.0\% & 52/114 (45.6\%) & 0.0043 \\
\midrule
\textbf{Across seeds} & -- & \textbf{47.1\% $\pm$ 5.0\% (pop. 4.1\%)} & \textbf{3 of 3} \\
\bottomrule
\end{tabular}
\end{center}
\medskip

\subsection{Premise-Only Baseline -- Cameroon Pidgin English}
\vspace{0.8\baselineskip}

\begin{center}
\small
\begin{tabular}{cccc}
\toprule
\textbf{Seed} & \textbf{Mean accuracy $\pm$ SD across folds} & \textbf{Pooled accuracy} & \textbf{$p$ (vs 1/3)} \\
\midrule
42 & 37.7\% $\pm$ 2.0\% & 43/114 (37.7\%) & 0.1851 \\
123 & 38.7\% $\pm$ 8.5\% & 44/114 (38.6\%) & 0.1376 \\
2024 & 35.1\% $\pm$ 7.6\% & 40/114 (35.1\%) & 0.3791 \\
\midrule
\textbf{Across seeds} & -- & \textbf{37.1\% $\pm$ 1.8\% (pop. 1.5\%)} & \textbf{0 of 3} \\
\bottomrule
\end{tabular}
\end{center}
\medskip
\noindent\textit{The model sees only the premise. With balanced labels, a premise alone should carry little label information, so results near 33.3\% are expected.}
\medskip

\Needspace{12\baselineskip}

\subsection{Premise-Only Baseline -- Camfranglais}

\begin{center}
\small
\begin{tabular}{cccc}
\toprule
\textbf{Seed} & \textbf{Mean accuracy $\pm$ SD across folds} & \textbf{Pooled accuracy} & \textbf{$p$ (vs 1/3)} \\
\midrule
42 & 36.8\% $\pm$ 2.2\% & 42/114 (36.8\%) & 0.2417 \\
123 & 44.7\% $\pm$ 7.9\% & 51/114 (44.7\%) & 0.0074 \\
2024 & 33.4\% $\pm$ 4.6\% & 38/114 (33.3\%) & 0.5352 \\
\midrule
\textbf{Across seeds} & -- & \textbf{38.3\% $\pm$ 5.8\% (pop. 4.8\%)} & \textbf{1 of 3} \\
\bottomrule
\end{tabular}
\end{center}
\medskip
\noindent\textit{Rounding note. The reported mean $\pm$ SD summaries are computed from the underlying fold-level and item-level outputs before rounding. The one-decimal seed-level values in these tables are rounded for presentation and may not reproduce the displayed summary statistics exactly when recomputed from the rounded values.}

\section{ASR / WER Diagnostic Record}
\label{app:asr}
This appendix preserves the detailed Whisper diagnostics for all three models (Base, Small, Large-v3) under three language conditions (automatic language identification, forced English, forced French) for both varieties: 18 blocks of 114 recordings, 2,052 transcripts. Raw WER is computed on the text as written (only Whisper language tags and extra whitespace removed). Normalized scoring uses 1,055 reference words per block for Cameroon Pidgin English and 1,096 for Camfranglais. All other values use the normalization rule in Appendix~\ref{app:asr_norm}; Appendix~\ref{app:asr_ci} gives the range of each WER, and Appendix~\ref{app:asr_script} gives the number of transcripts with non-Latin script and the number left empty after normalization.

\subsection{Text Normalization and Metric Definitions}
\label{app:asr_norm}
\noindent\textbf{Normalization.} Normalization is applied identically to the reference and the Whisper output: (1) delete Whisper language tags of the form \texttt{<|xx|>}; (2) convert to Unicode NFC and lowercase; (3) convert curly apostrophes to straight ones; (4) keep apostrophes, because they belong to French words (j'ai, l'école); (5) replace every other character that is not a letter, a digit or a combining mark with a space; (6) collapse whitespace. All letters, including accented letters, are kept. Example: ``Je suis au carrefour comme ça.'' becomes ``je suis au carrefour comme ça''.

\smallskip
\noindent\textbf{Metrics.} Raw WER is computed on the unnormalized strings split on whitespace (case and punctuation kept; only Whisper tags and extra whitespace are removed). Normalized WER is computed on the normalized strings and is the value reported in the main paper. MER, WIL, and WIP are the standard match-error, word-information-lost and word-information-preserved measures. CER is the character error rate on the normalized text. S, D and I are substitutions, deletions and insertions from a minimum-edit-distance alignment, and Correct is the number of matched words. Distinct language tags is the number of different languages Whisper predicted across the 114 clips under automatic language identification (one when the language is forced).

\subsection{Final Primary ASR Record -- Cameroon Pidgin English (automatic language ID)}
\begin{center}
\small
\begin{tabular}{lccc}
\toprule
\textbf{Metric} & \textbf{Base} & \textbf{Small} & \textbf{Large-v3} \\
\midrule
Raw WER & 94.6\% & 98.3\% & 84.4\% \\
Normalized WER & 91.8\% & 95.7\% & 80.6\% \\
MER & 85.7\% & 90.5\% & 77.2\% \\
WIL & 97.2\% & 98.7\% & 93.3\% \\
WIP & 2.8\% & 1.3\% & 6.7\% \\
CER (normalized) & 61.9\% & 70.2\% & 45.6\% \\
Reference words & 1,055 & 1,055 & 1,055 \\
Hypothesis words & 882 & 801 & 898 \\
Correct & 161 & 106 & 251 \\
Substitutions & 647 & 634 & 601 \\
Deletions & 247 & 315 & 203 \\
Insertions & 74 & 61 & 46 \\
Distinct language tags & 20 & 11 & 9 \\
\bottomrule
\end{tabular}
\end{center}

\subsection{Cameroon Pidgin -- Forced-English Diagnostic}
\begin{center}
\small
\begin{tabular}{lccc}
\toprule
\textbf{Metric} & \textbf{Base} & \textbf{Small} & \textbf{Large-v3} \\
\midrule
Raw WER & 84.0\% & 79.2\% & 72.0\% \\
Normalized WER & 80.3\% & 74.1\% & 66.6\% \\
MER & 76.2\% & 71.2\% & 64.2\% \\
WIL & 93.1\% & 90.3\% & 85.4\% \\
WIP & 6.9\% & 9.7\% & 14.6\% \\
CER (normalized) & 48.1\% & 39.0\% & 32.3\% \\
Reference words & 1,055 & 1,055 & 1,055 \\
Hypothesis words & 961 & 972 & 1,000 \\
Correct & 265 & 316 & 392 \\
Substitutions & 639 & 613 & 568 \\
Deletions & 151 & 126 & 95 \\
Insertions & 57 & 43 & 40 \\
\bottomrule
\end{tabular}
\end{center}

\subsection{Cameroon Pidgin -- Forced-French Diagnostic}
\begin{center}
\small
\begin{tabular}{lccc}
\toprule
\textbf{Metric} & \textbf{Base} & \textbf{Small} & \textbf{Large-v3} \\
\midrule
Raw WER & 137.9\% & 101.6\% & 101.3\% \\
Normalized WER & 136.2\% & 98.3\% & 98.3\% \\
MER & 95.9\% & 91.7\% & 89.8\% \\
WIL & 99.7\% & 99.1\% & 98.7\% \\
WIP & 0.3\% & 0.9\% & 1.3\% \\
CER (normalized) & 116.3\% & 72.6\% & 68.0\% \\
Reference words & 1,055 & 1,055 & 1,055 \\
Hypothesis words & 1,331 & 954 & 1,024 \\
Correct & 61 & 94 & 118 \\
Substitutions & 827 & 784 & 806 \\
Deletions & 167 & 177 & 131 \\
Insertions & 443 & 76 & 100 \\
\bottomrule
\end{tabular}
\end{center}
\medskip
\noindent\textit{Forced English lowers the normalized WER at every model size compared with automatic language identification (for Large-v3, 66.6\% against 80.6\%), whereas forced French raises it at every size. Forcing French is a mismatched-language control for Pidgin. Base inserts many words (443 insertions), which pushes its WER above 100\%. Small and Large-v3 reach the same normalized WER (98.3\%) by coincidence: each makes 1,037 errors against 1,055 reference words, but with a different mix (784 substitutions, 177 deletions and 76 insertions for Small; 806, 131 and 100 for Large-v3).}
\medskip

\subsection{Final Primary ASR Record -- Camfranglais (automatic language ID)}
\begin{center}
\small
\begin{tabular}{lccc}
\toprule
\textbf{Metric} & \textbf{Base} & \textbf{Small} & \textbf{Large-v3} \\
\midrule
Raw WER & 94.3\% & 82.1\% & 71.0\% \\
Normalized WER & 90.1\% & 77.1\% & 64.9\% \\
MER & 81.7\% & 73.7\% & 63.1\% \\
WIL & 95.5\% & 91.0\% & 84.4\% \\
WIP & 4.5\% & 9.0\% & 15.6\% \\
CER (normalized) & 64.5\% & 52.5\% & 33.4\% \\
Reference words & 1,096 & 1,096 & 1,096 \\
Hypothesis words & 1,002 & 913 & 1,011 \\
Correct & 221 & 301 & 416 \\
Substitutions & 668 & 562 & 564 \\
Deletions & 207 & 233 & 116 \\
Insertions & 113 & 50 & 31 \\
Distinct language tags & 23 & 17 & 17 \\
\bottomrule
\end{tabular}
\end{center}

\noindent\textit{Automatic language identification is the primary condition. Normalized WER falls from 90.1\% (Base) to 64.9\% (Large-v3), and correct words rise from 221 to 416 of 1,096 reference words. Large-v3 also makes the fewest deletions (116) and insertions (31). Base produces 23 distinct language tags, against 17 for Small and Large-v3.}

\subsection{Camfranglais -- Forced-French Diagnostic}
\vspace{1.0\baselineskip}

\begin{center}
\small
\begin{tabular}{lccc}
\toprule
\textbf{Metric} & \textbf{Base} & \textbf{Small} & \textbf{Large-v3} \\
\midrule
Raw WER & 85.7\% & 69.1\% & 55.9\% \\
Normalized WER & 79.3\% & 61.8\% & 45.5\% \\
MER & 70.9\% & 59.5\% & 44.2\% \\
WIL & 89.7\% & 80.8\% & 66.2\% \\
WIP & 10.3\% & 19.2\% & 33.8\% \\
CER (normalized) & 54.8\% & 39.3\% & 26.3\% \\
Reference words & 1,096 & 1,096 & 1,096 \\
Hypothesis words & 1,119 & 1,006 & 1,071 \\
Correct & 356 & 460 & 630 \\
Substitutions & 634 & 505 & 408 \\
Deletions & 106 & 131 & 58 \\
Insertions & 129 & 41 & 33 \\
\bottomrule
\end{tabular}
\end{center}

\vspace{1.0\baselineskip}
\noindent\textit{Normalized WER falls from 79.3\% (Base) to 45.5\% (Large-v3), and correct words rise from 356 to 630 of 1,096 reference words. Large-v3 also makes the fewest insertions (33) and deletions (58). This run is diagnostic: forced French requires knowing the language in advance, and the recordings are Camfranglais rather than standard French.}

\vspace{1.0\baselineskip}

\subsection{Camfranglais -- Forced-English Diagnostic}
\vspace{1.0\baselineskip}

\begin{center}
\small
\begin{tabular}{lccc}
\toprule
\textbf{Metric} & \textbf{Base} & \textbf{Small} & \textbf{Large-v3} \\
\midrule
Raw WER & 101.0\% & 100.0\% & 95.6\% \\
Normalized WER & 100.2\% & 98.2\% & 93.9\% \\
MER & 97.4\% & 92.7\% & 89.3\% \\
WIL & 99.9\% & 99.3\% & 98.6\% \\
WIP & 0.1\% & 0.7\% & 1.4\% \\
CER (normalized) & 76.0\% & 68.4\% & 62.4\% \\
Reference words & 1,096 & 1,096 & 1,096 \\
Hypothesis words & 775 & 979 & 993 \\
Correct & 29 & 85 & 123 \\
Substitutions & 715 & 829 & 814 \\
Deletions & 352 & 182 & 159 \\
Insertions & 31 & 65 & 56 \\
\bottomrule
\end{tabular}
\end{center}
\medskip

\vspace{1.0\baselineskip}
\noindent\textit{Forced English is a mismatched-language control for Camfranglais, which is largely French-based. In the forced-language tables (C.6 and C.7), the language tag is fixed, so the ``distinct language tags'' row is omitted. For all three models, forced French gives the lowest normalized WER, automatic identification is in the middle, and forced English is the highest (for Large-v3, 45.5\%, 64.9\%, and 93.9\%). For Camfranglais, WER falls with model size in every condition. Under forced English, the normalized WER of Base is slightly above 100\% (100.2\%) because its total errors, including insertions, exceed the number of reference words, and Large-v3 recovers only 123 of the 1,096 reference words correctly, against 416 under automatic identification and 630 under forced French.}
\medskip

\clearpage
\subsection{Representative Transcripts (Whisper Large-v3)}
\label{app:asr_examples}

\begin{center}
\small
\begin{tabular}{l>{\raggedright\arraybackslash}p{0.2\textwidth}>{\raggedright\arraybackslash}p{0.2\textwidth}>{\raggedright\arraybackslash}p{0.2\textwidth}>{\raggedright\arraybackslash}p{0.18\textwidth}}
\toprule
\textbf{Clip} & \textbf{Reference} & \textbf{Automatic (detected language)} & \textbf{Forced language} & \textbf{What it shows} \\
\midrule
\texttt{wes\_00018} & price fo petrol don climb again. & price for petrodon climb again (en) & same & Pidgin recognized as English; WER 50\% \\
\midrule
\texttt{wes\_00103} & president fo njangi house broke bank di end of year ma own share fo de money no comot. & president Funchangi has broke bank for this end of year. Ma own share for the Molino Commod. (yo) & President Funjangi has broke bank for this end of year. Maun She for the Molino Commode. & Local words replaced by wrong ones (njangi $\rightarrow$ Funchangi); WER 56\% $\rightarrow$ 72\% \\
\midrule
\texttt{wes\_00008} & e ask me how old a dey. & \includegraphics[width=\linewidth,height=0.5cm,keepaspectratio]{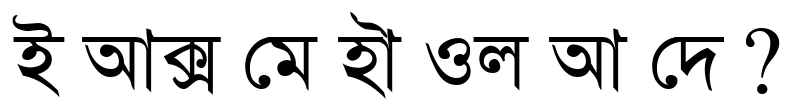}\newline (yo) & He asked me how old I'd be. & Non-Latin script for a Latin-script reference; WER 100\% $\rightarrow$ 57\% \\
\midrule
\texttt{wes\_00086} & ma phone e screen crack crack very badly. & \includegraphics[width=\linewidth,height=0.5cm,keepaspectratio]{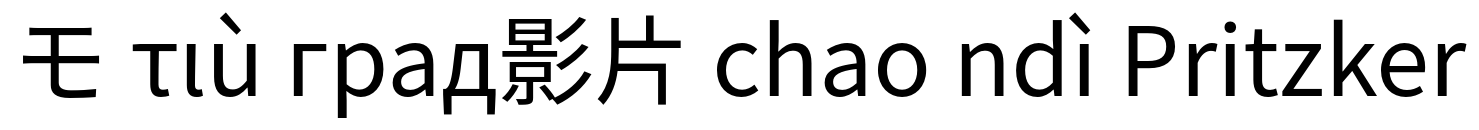}\newline (sn) & My phone is screen crack crack very badly. & Four scripts in one transcript; forcing English fixes it; WER 100\% $\rightarrow$ 25\% \\
\midrule
\texttt{camf\_00019} & je suis moi peace. & Je suis moi, peace ! (fr) & same & Correct; WER 0\% \\
\midrule
\texttt{camf\_00018} & le prix du carburant a encore gonfle. & Lo pri di kabir an a anko gonfle. (ht) & Le prix du carburant a encore gonflé. & Haitian Creole tag; forcing French improves WER 86\% $\rightarrow$ 14\% \\
\midrule
\texttt{camf\_00023} & le grand vent a arraché les tôles. & \includegraphics[width=\linewidth,height=0.5cm,keepaspectratio]{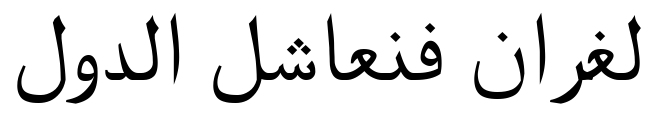}\newline (ar) & Le grand vent a acheté le tout. & Arabic script for French speech; WER 100\% $\rightarrow$ 43\% \\
\midrule
\texttt{camf\_00009} & Paul a travel pour Doual la semaine passée. & Paul a travaillé pour Douala la semaine passée. (fr) & same & Local spelling ``Doual'' and the English word ``travel'' replaced; WER 25\% \\
\bottomrule
\end{tabular}
\end{center}
\medskip

\noindent\textit{Examples are copied unchanged from the saved transcripts and illustrate language misidentification, wrong script, wrong local words, and spelling mismatch. Clip-level WER is normalized; where two values are shown, the first is for automatic decoding and the second for forced decoding (English for Pidgin, French for Camfranglais).\\
Because neither variety has a standard spelling, a phonetic transcription counts as an error even though the sounds were recovered. Under Large-v3, no Pidgin recording is error-free in any condition; for Camfranglais, 8 of 114 recordings are error-free under automatic identification, 3 under forced English and 10 under forced French. The row-level record for all 114 recordings, every model and condition, is included in the release.}

\subsection{Language Identification Distribution Under Automatic Decoding}
\label{app:asr_lid}

\vspace{1.0\baselineskip}
\begin{center}
\small
\begin{tabular}{llcll}
\toprule
\textbf{Variety} & \textbf{Model} & \textbf{Distinct tags} & \textbf{Matching-language tag (count)} & \textbf{Most frequent tags (of 114)} \\
\midrule
Cameroon Pidgin & Base & 20 & en 16 & yo 31, en 16, sw 15, de 12 \\
Cameroon Pidgin & Small & 11 & en 3 & yo 51, sw 35, fr 7, sn 6 \\
Cameroon Pidgin & Large-v3 & 9 & en 9 & yo 53, sw 40, en 9, sn 4 \\
\midrule
Camfranglais & Base & 23 & fr 51 & fr 51, ar 11, es 6, de 6 \\
Camfranglais & Small & 17 & fr 58 & fr 58, yo 13, ht 12, ar 7 \\
Camfranglais & Large-v3 & 17 & fr 54 & fr 54, ht 24, sw 9, yo 8 \\
\bottomrule
\end{tabular}
\end{center}
\medskip

\noindent\textit{Whisper rarely identifies Cameroon Pidgin as English. Under automatic decoding, Large-v3 tags it as Yoruba or Swahili for 93 of 114 recordings and as English for only 9. Camfranglais is most often identified as French, but 24 recordings are tagged as Haitian Creole under Large-v3.\\
Tags are ISO 639-1 codes (yo = Yoruba, en = English, sw = Swahili, de = German, fr = French, ar = Arabic, es = Spanish, ht = Haitian Creole, sn = Shona).\\
Clips whose automatic transcript contains non-Latin script: Cameroon Pidgin (Base 26, Small 34, Large-v3 14), Camfranglais (Base 24, Small 26, Large-v3 7). The complete tag counts for all 18 blocks are in \texttt{asr\_summary.csv} in the release.}

\subsection{Ranges of the WER}
\label{app:asr_ci}
The WER is computed on 114 recordings, so it would change if different recordings were used. Each range shows how much it could change. Overlapping ranges neither establish nor rule out a difference, because every model is scored on the same recordings. Ranges are 95\% bootstrap intervals (recordings resampled 2,000 times, seed 0).

\begin{center}
\small
\begin{tabular}{llccc}
\toprule
\textbf{Variety} & \textbf{Condition} & \textbf{Base} & \textbf{Small} & \textbf{Large-v3} \\
\midrule
Cameroon Pidgin & Automatic & 91.8\% (85.0--101.4) & 95.7\% (91.6--100.2) & 80.6\% (75.6--85.7) \\
& Forced English & 80.3\% (76.6--84.2) & 74.1\% (70.5--78.0) & 66.6\% (62.7--70.6) \\
& Forced French & 136.2\% (117.0--160.9) & 98.3\% (94.1--102.6) & 98.3\% (93.5--102.7) \\
\midrule
Camfranglais & Automatic & 90.1\% (82.4--99.2) & 77.1\% (70.4--83.8) & 64.9\% (58.3--71.6) \\
& Forced English & 100.2\% (98.4--102.4) & 98.2\% (94.8--101.5) & 93.9\% (89.9--97.6) \\
& Forced French & 79.3\% (71.5--87.8) & 61.8\% (56.0--67.3) & 45.5\% (41.0--50.4) \\
\bottomrule
\end{tabular}
\end{center}

\subsection{Script and Empty-Transcript Checks}
\label{app:asr_script}
For each variety and condition, the number of the 114 transcripts that contain letters outside the Latin script (Non-Latin) and the number that are empty after normalization (Empty), listed for Base / Small / Large-v3.

\begin{center}
\small
\begin{tabular}{llcc}
\toprule
\textbf{Variety} & \textbf{Condition} & \textbf{Non-Latin (Base / Small / Large-v3)} & \textbf{Empty (Base / Small / Large-v3)} \\
\midrule
Cameroon Pidgin & Automatic & 26 / 34 / 14 & 1 / 0 / 0 \\
& Forced English & 4 / 0 / 2 & 0 / 0 / 0 \\
& Forced French & 27 / 14 / 1 & 0 / 0 / 1 \\
\midrule
Camfranglais & Automatic & 24 / 26 / 7 & 0 / 0 / 0 \\
& Forced English & 10 / 5 / 2 & 1 / 1 / 0 \\
& Forced French & 11 / 2 / 0 & 0 / 0 / 0 \\
\bottomrule
\end{tabular}
\end{center}

\section{Cross-Variety Transfer Results}
\label{app:transfer}

\subsection{Protocols}
\noindent\textbf{Disjoint-item transfer (main result).} Using the shared item folds (Appendix~\ref{app:folds}), fold $k$ trains on source-variety items outside fold $k$ (91--92 items) and tests on target-variety items in fold $k$ (22--23 items). The five folds together give 114 test predictions per seed. A test item's counterpart (the same scenario in the other variety) is never in training.

\noindent\textbf{Full-source transfer (counterpart-leak audit).} The model is trained on all 114 source items and tested on all 114 target items. Every test item's counterpart is therefore in the training set (114/114). This protocol is reported only as an audit, and the numbers must not be read as cross-variety transfer.

\noindent\textbf{Shared settings.} The same XLM-RoBERTa configuration was used in both directions (Appendix~\ref{app:config_xlmr}). The three seeds are repeated runs on the same items and are therefore not independent test sets. Mean $\pm$ SD across seeds uses pooled accuracy and the sample SD (population SD in brackets). The last column gives the one-sided exact binomial $p$-value against 1/3 (47/114 is the minimum for $p < 0.05$); in the ``Across seeds'' rows it counts the seeds with $p < 0.05$. Chance level is 33.3\%.

\vspace{0.6\baselineskip}
\subsection{Disjoint-Item Transfer, Full Input}
\label{app:tr_disjoint}
\vspace{0.6\baselineskip}
\begin{center}
\small
\begin{tabular}{lcccc}
\toprule
\textbf{Training $\rightarrow$ Test} & \textbf{Seed} & \textbf{Mean $\pm$ SD across folds} & \textbf{Pooled accuracy} & \textbf{$p$} \\
\midrule
Pidgin $\rightarrow$ Camfranglais & 42 & 37.7\% $\pm$ 10.3\% & 43/114 (37.7\%) & 0.1851 \\
& 123 & 43.8\% $\pm$ 6.4\% & 50/114 (43.9\%) & 0.0123 \\
& 2024 & 38.6\% $\pm$ 3.8\% & 44/114 (38.6\%) & 0.1376 \\
\midrule
\textbf{Across seeds} & & -- & \textbf{40.1\% $\pm$ 3.3\% (pop. 2.7\%)} & \textbf{1 of 3} \\
\midrule
Camfranglais $\rightarrow$ Pidgin & 42 & 42.1\% $\pm$ 4.4\% & 48/114 (42.1\%) & 0.0312 \\
& 123 & 45.5\% $\pm$ 12.2\% & 52/114 (45.6\%) & 0.0043 \\
& 2024 & 49.0\% $\pm$ 7.7\% & 56/114 (49.1\%) & 0.0004 \\
\midrule
\textbf{Across seeds} & & -- & \textbf{45.6\% $\pm$ 3.5\% (pop. 2.9\%)} & \textbf{3 of 3} \\
\bottomrule
\end{tabular}
\end{center}

\vspace{0.6\baselineskip}
\subsection{Disjoint-Item Transfer, Hypothesis-Only Input}
\vspace{0.6\baselineskip}
\begin{center}
\small
\begin{tabular}{lcccc}
\toprule
\textbf{Training $\rightarrow$ Test} & \textbf{Seed} & \textbf{Mean $\pm$ SD across folds} & \textbf{Pooled accuracy} & \textbf{$p$} \\
\midrule
Pidgin $\rightarrow$ Camfranglais & 42 & 46.5\% $\pm$ 6.5\% & 53/114 (46.5\%) & 0.0024 \\
& 123 & 48.1\% $\pm$ 8.1\% & 55/114 (48.2\%) & 0.0007 \\
& 2024 & 37.7\% $\pm$ 6.2\% & 43/114 (37.7\%) & 0.1851 \\
\midrule
\textbf{Across seeds} & & -- & \textbf{44.2\% $\pm$ 5.6\% (pop. 4.6\%)} & \textbf{2 of 3} \\
\midrule
Camfranglais $\rightarrow$ Pidgin & 42 & 46.5\% $\pm$ 5.2\% & 53/114 (46.5\%) & 0.0024 \\
& 123 & 48.2\% $\pm$ 7.3\% & 55/114 (48.2\%) & 0.0007 \\
& 2024 & 51.8\% $\pm$ 5.0\% & 59/114 (51.8\%) & $< 0.0001$ \\
\midrule
\textbf{Across seeds} & & -- & \textbf{48.8\% $\pm$ 2.7\% (pop. 2.2\%)} & \textbf{3 of 3} \\
\bottomrule
\end{tabular}
\end{center}
\medskip
\noindent\textit{The hypothesis-only means (44.2\% and 48.8\%) are higher than the full-input transfer means (40.1\% and 45.6\%). Full-input transfer is therefore not shown to use the premise (see Appendix~\ref{app:mcnemar}).}

\subsection{Full-Source Transfer (Counterpart-Leak Audit, Not a Transfer Result)}
\label{app:tr_leak}

\vspace{1.0\baselineskip}
\begin{center}
\small
\begin{tabular}{lcccc}
\toprule
\textbf{Training $\rightarrow$ Test} & \textbf{Seed} & \textbf{Mean $\pm$ SD across folds} & \textbf{Accuracy (all 114 test items)} & \textbf{$p$} \\
\midrule
Pidgin $\rightarrow$ Camfranglais & 42 & -- & 72/114 (63.2\%) & $< 0.0001$ \\
& 123 & -- & 79/114 (69.3\%) & $< 0.0001$ \\
& 2024 & -- & 67/114 (58.8\%) & $< 0.0001$ \\
\midrule
\textbf{Across seeds} & & -- & \textbf{63.7\% $\pm$ 5.3\% (pop. 4.3\%)} & \textbf{3 of 3} \\
\midrule
Camfranglais $\rightarrow$ Pidgin & 42 & -- & 79/114 (69.3\%) & $< 0.0001$ \\
& 123 & -- & 67/114 (58.8\%) & $< 0.0001$ \\
& 2024 & -- & 61/114 (53.5\%) & $< 0.0001$ \\
\midrule
\textbf{Across seeds} & & -- & \textbf{60.5\% $\pm$ 8.0\% (pop. 6.6\%)} & \textbf{3 of 3} \\
\bottomrule
\end{tabular}
\end{center}
\medskip
\noindent\textit{There are no folds in this protocol (one training run per seed); the ``Mean $\pm$ SD across folds'' column is empty. Compared with disjoint-item transfer (Appendix~\ref{app:tr_disjoint}), accuracy is about 24 points higher for Pidgin $\rightarrow$ Camfranglais and 15 points higher for Camfranglais $\rightarrow$ Pidgin. That gap reflects counterpart leakage together with the larger training set (114 instead of 91--92 items).}

\vspace{0.8\baselineskip}
\subsection{Side-by-Side Comparison}
\label{app:tr_side}

\begin{center}
\setlength{\tabcolsep}{3pt}
\begin{tabular*}{\linewidth}{@{\extracolsep{\fill}}lcccc}
\toprule
\textbf{Direction}
& \shortstack{\textbf{Disjoint,}\\\textbf{full input}}
& \shortstack{\textbf{Disjoint,}\\\textbf{hypothesis-only}}
& \shortstack{\textbf{Within-variety}\\\textbf{(target)}}
& \shortstack{\textbf{Full-source}\\\textbf{(counterpart leak)}} \\
\midrule
Pidgin $\rightarrow$ Camfranglais & 40.1\% $\pm$ 3.3\% & 44.2\% $\pm$ 5.6\% & 46.2\% $\pm$ 1.8\% & 63.7\% $\pm$ 5.3\% \\
Camfranglais $\rightarrow$ Pidgin & 45.6\% $\pm$ 3.5\% & 48.8\% $\pm$ 2.7\% & 44.7\% $\pm$ 2.3\% & 60.5\% $\pm$ 8.0\% \\
\bottomrule
\end{tabular*}
\end{center}

\medskip
\noindent\textit{Note. Mean $\pm$ sample SD over seeds 42, 123 and 2024. ``Within-variety (target)'' is the five-fold accuracy of a model trained and tested on the target variety (Camfranglais in the first row, Pidgin in the second). Per seed, letting counterparts into training raises the number of correct predictions by 29, 29 and 23 (Pidgin $\rightarrow$ Camfranglais) and by 31, 15 and 5 (Camfranglais $\rightarrow$ Pidgin) relative to the disjoint run with the same seed, an average gain of 23.7 and 14.9 points, positive in all six seed pairs. Disjoint transfer is 6.1 points below within-variety accuracy on Camfranglais, and 0.9 points above it on Pidgin; the 0.9-point difference on Pidgin is within the seed-to-seed variation, whereas on Camfranglais all three transfer seeds (37.7\%, 43.9\%, 38.6\%) lie below all three within-variety seeds (45.6\%, 44.7\%, 48.2\%).}

\section{Epoch Sensitivity}
\label{app:epoch}

\subsection*{Setup}
\noindent The model, data, folds, and seeds are identical to the main within-variety runs (native text, full input); only the number of epochs changes (5, 10, 15, and 20). The 15-epoch values are the main runs, reused. Each epoch is 23 optimizer steps (115 to 460 in total). The two varieties are reported separately (Tables~E.1 and~E.2). Hypothesis-only and transfer runs used 15 epochs only.

\noindent These runs are a sensitivity analysis on the same data, folds, and seeds. Because there is no separate development set, the main accuracies may be slightly optimistic. Accuracy rises most from 5 to 10 epochs ($+7.0$ points for Pidgin, $+5.5$ for Camfranglais), and each further step adds at most $2.3$ points. Fifteen epochs is not the highest-scoring setting for either variety, but from 15 to 20 epochs the mean changes by only $+1.2$ (Pidgin) and $+2.3$ points (Camfranglais), less than the across-seed SD at 20 epochs ($5.0$ and $4.1$). We therefore keep 15 epochs as the main setting.

\medskip
\noindent\textbf{Table E.1: Accuracy by number of epochs (pooled accuracy, mean $\pm$ sample SD over seeds 42, 123, 2024)}

\medskip
\begin{center}
\small
\begin{tabular}{lcccc}
\toprule
\textbf{Variety} & \textbf{Epochs} & \textbf{Mean accuracy $\pm$ SD (pop.)} & \textbf{Change from 15 epochs (points)} & \textbf{Seeds $p < 0.05$} \\
\midrule
Cameroon Pidgin & 5 & 35.7\% $\pm$ 3.1\% (pop. 2.5\%) & $-9.1$ & 0 of 3 \\
& 10 & 42.7\% $\pm$ 3.5\% (pop. 2.9\%) & $-2.0$ & 2 of 3 \\
& 15 (main) & 44.7\% $\pm$ 2.3\% (pop. 1.9\%) & -- & 3 of 3 \\
& 20 & 45.9\% $\pm$ 5.0\% (pop. 4.1\%) & $+1.2$ & 2 of 3 \\
\midrule
Camfranglais & 5 & 38.9\% $\pm$ 2.2\% (pop. 1.8\%) & $-7.3$ & 1 of 3 \\
& 10 & 44.4\% $\pm$ 6.2\% (pop. 5.1\%) & $-1.8$ & 2 of 3 \\
& 15 (main) & 46.2\% $\pm$ 1.8\% (pop. 1.5\%) & -- & 3 of 3 \\
& 20 & 48.5\% $\pm$ 4.1\% (pop. 3.4\%) & $+2.3$ & 3 of 3 \\
\bottomrule
\end{tabular}
\end{center}

\medskip
\noindent\textit{Note. Five epochs give the lowest accuracy for both varieties (near chance for Pidgin). Mean accuracy rises from 5 to 15 epochs in both varieties, and 15 epochs is the earliest setting at which all three seeds are above chance for both of them; for Pidgin, 20 epochs is back to two of three.}

\vspace{1.0\baselineskip}
\noindent\textbf{Table E.2: Per-seed results by number of epochs}
\vspace{1.0\baselineskip}
\begin{center}
\small
\begin{tabular}{lccccc}
\toprule
\textbf{Variety} & \textbf{Epochs} & \textbf{Seed 42} & \textbf{Seed 123} & \textbf{Seed 2024} & \textbf{Seeds $p < 0.05$} \\
\midrule
Cameroon Pidgin & 5 & 37 (0.6132) & 44 (0.1376) & 41 (0.3069) & 0 of 3 \\
& 10 & 45 (0.0994) & 53 (0.0024) & 48 (0.0312) & 2 of 3 \\
& 15 (main) & 52 (0.0043) & 53 (0.0024) & 48 (0.0312) & 3 of 3 \\
& 20 & 54 (0.0013) & 46 (0.0696) & 57 (0.0002) & 2 of 3 \\
\midrule
Camfranglais & 5 & 44 (0.1376) & 47 (0.0473) & 42 (0.2417) & 1 of 3 \\
& 10 & 43 (0.1851) & 52 (0.0043) & 57 (0.0002) & 2 of 3 \\
& 15 (main) & 52 (0.0043) & 51 (0.0074) & 55 (0.0007) & 3 of 3 \\
& 20 & 50 (0.0123) & 57 (0.0002) & 59 ($< 0.0001$) & 3 of 3 \\
\bottomrule
\end{tabular}
\end{center}

\vspace{1.0\baselineskip}
\noindent\textit{Correct out of 114; one-sided exact binomial $p$-values against 1/3 are in brackets. 47/114 is the minimum for $p < 0.05$.}

\section{Statistical Tests}
\label{app:stats}

\subsection{One-Sided Binomial Tests Against Chance}
For each condition and seed, we test whether the pooled accuracy over 114 predictions exceeds 1/3 (one-sided exact binomial). The smallest passing score is 47/114 (41.2\%, $p = 0.047$); 46/114 gives $p = 0.070$. A single run's 95\% interval is roughly $\pm 9$ points around its accuracy, so differences of a few points between conditions are within noise. Seeds are repeated runs on the same items, so ``seeds above chance'' describes consistency across runs, not independent replication.

\vspace{1.0\baselineskip}

\begin{center}
\small
\begin{tabular}{lcccc}
\toprule
\textbf{Condition} & \textbf{Seed 42} & \textbf{Seed 123} & \textbf{Seed 2024} & \textbf{Seeds $p < 0.05$} \\
\midrule
Pidgin, within, full & 52 (0.0043) & 53 (0.0024) & 48 (0.0312) & 3 of 3 \\
Camfranglais, within, full & 52 (0.0043) & 51 (0.0074) & 55 (0.0007) & 3 of 3 \\
Pidgin, within, hypothesis-only & 46 (0.0696) & 50 (0.0123) & 47 (0.0473) & 2 of 3 \\
Camfranglais, within, hypothesis-only & 60 ($< 0.0001$) & 49 (0.0199) & 52 (0.0043) & 3 of 3 \\
Pidgin, within, premise-only & 43 (0.1851) & 44 (0.1376) & 40 (0.3791) & 0 of 3 \\
Camfranglais, within, premise-only & 42 (0.2417) & 51 (0.0074) & 38 (0.5352) & 1 of 3 \\
Pidgin gloss & 73 ($< 0.0001$) & 64 ($< 0.0001$) & 72 ($< 0.0001$) & 3 of 3 \\
Camfranglais gloss & 73 ($< 0.0001$) & 64 ($< 0.0001$) & 72 ($< 0.0001$) & 3 of 3 \\
Pidgin $\rightarrow$ Camfranglais, disjoint, full & 43 (0.1851) & 50 (0.0123) & 44 (0.1376) & 1 of 3 \\
Camfranglais $\rightarrow$ Pidgin, disjoint, full & 48 (0.0312) & 52 (0.0043) & 56 (0.0004) & 3 of 3 \\
Pidgin $\rightarrow$ Camfranglais, disjoint, hyp.-only & 53 (0.0024) & 55 (0.0007) & 43 (0.1851) & 2 of 3 \\
Camfranglais $\rightarrow$ Pidgin, disjoint, hyp.-only & 53 (0.0024) & 55 (0.0007) & 59 ($< 0.0001$) & 3 of 3 \\
Pidgin $\rightarrow$ Camfranglais, full-source (leak) & 72 ($< 0.0001$) & 79 ($< 0.0001$) & 67 ($< 0.0001$) & 3 of 3 \\
Camfranglais $\rightarrow$ Pidgin, full-source (leak) & 79 ($< 0.0001$) & 67 ($< 0.0001$) & 61 ($< 0.0001$) & 3 of 3 \\
\bottomrule
\end{tabular}
\end{center}

\vspace{1.0\baselineskip}
\noindent\textit{Cells give the correct out of 114 and the one-sided $p$-value. The two gloss rows are identical because the glosses, labels, and folds are shared (Section~\ref{subsec:setup}). Epoch-sensitivity tests are in Appendix~\ref{app:epoch}.}
\medskip

\needspace{10\baselineskip}
\subsection{Full Input Versus Hypothesis-Only (Exact McNemar Tests)}
\label{app:mcnemar}
For each seed and setting, we compare item-level correctness of the full-input and hypothesis-only models on the same 114 test items. ``Full only'' counts items the full-input model got right and the hypothesis-only model got wrong, and ``Hyp. only'' the reverse. The $p$-value is the exact two-sided McNemar (binomial) test on the discordant items.

\vspace{1.0\baselineskip}

\begin{center}
\small
\begin{tabular}{lcccccc}
\toprule
\textbf{Setting} & \textbf{Seed} & \textbf{Full correct} & \textbf{Hyp. correct} & \textbf{Full only} & \textbf{Hyp. only} & \textbf{$p$} \\
\midrule
Pidgin, within & 42 & 52 & 46 & 25 & 19 & 0.451 \\
& 123 & 53 & 50 & 24 & 21 & 0.766 \\
& 2024 & 48 & 47 & 20 & 19 & 1.000 \\
\midrule
Camfranglais, within & 42 & 52 & 60 & 20 & 28 & 0.312 \\
& 123 & 51 & 49 & 21 & 19 & 0.875 \\
& 2024 & 55 & 52 & 18 & 15 & 0.728 \\
\midrule
Pidgin $\rightarrow$ Camfranglais, disjoint & 42 & 43 & 53 & 21 & 31 & 0.212 \\
& 123 & 50 & 55 & 21 & 26 & 0.560 \\
& 2024 & 44 & 43 & 22 & 21 & 1.000 \\
\midrule
Camfranglais $\rightarrow$ Pidgin, disjoint & 42 & 48 & 53 & 21 & 26 & 0.560 \\
& 123 & 52 & 55 & 22 & 25 & 0.771 \\
& 2024 & 56 & 59 & 23 & 26 & 0.775 \\
\bottomrule
\end{tabular}
\end{center}

\vspace{1.0\baselineskip}
\noindent\textit{Twelve comparisons were made and none has $p < 0.05$ (the smallest is 0.212). The full-input model scores higher in 6 of 12 comparisons and the hypothesis-only model in the other 6, and the sign changes between seeds within the same setting (Camfranglais within-variety and Pidgin $\rightarrow$ Camfranglais). The data therefore do not show that full input outperforms hypothesis-only. They also do not show that the two are equal: with 114 items and 33 to 52 discordant pairs per test, differences of several points are not statistically detectable.}
\medskip

\vspace{1.0\baselineskip}
\subsection{Mean Differences (Full Input Minus Hypothesis-Only, Pooled Accuracy)}

\vspace{1.0\baselineskip}
\begin{center}
\small
\begin{tabular}{lccc}
\toprule
\textbf{Setting} & \textbf{Full} & \textbf{Hypothesis-only} & \textbf{Difference} \\
\midrule
Pidgin, within & 44.7\% & 41.8\% & $+2.9$ \\
Camfranglais, within & 46.2\% & 47.1\% & $-0.9$ \\
Pidgin $\rightarrow$ Camfranglais, disjoint & 40.1\% & 44.2\% & $-4.1$ \\
Camfranglais $\rightarrow$ Pidgin, disjoint & 45.6\% & 48.8\% & $-3.2$ \\
\bottomrule
\end{tabular}
\end{center}

\vspace{1.0\baselineskip}
\noindent\textit{Differences are in percentage points, averaged over the three seeds. They are small, change sign across settings, and are within the seed-to-seed variation.}

\vspace{0.8\baselineskip}
\section{Cross-Validation Folds}
\label{app:folds}

\vspace{0.5\baselineskip}
\noindent\textbf{Table G.1: Shared item-level fold assignment (fold seed 42)}
\vspace{0.8\baselineskip}
\begin{center}
\small
\begin{tabular}{ccccc}
\toprule
\textbf{Fold} & \textbf{Items} & \textbf{Entailment} & \textbf{Contradiction} & \textbf{Neutral} \\
\midrule
0 & 23 & 8 & 7 & 8 \\
1 & 23 & 8 & 8 & 7 \\
2 & 23 & 8 & 8 & 7 \\
3 & 23 & 7 & 8 & 8 \\
4 & 22 & 7 & 7 & 8 \\
\midrule
\textbf{Total} & \textbf{114} & \textbf{38} & \textbf{38} & \textbf{38} \\
\bottomrule
\end{tabular}
\end{center}

\vspace{0.8\baselineskip}
\noindent\textit{One stratified five-fold assignment is defined on item numbers 1--114 and shared by both varieties: item $i$ is in the same fold in Pidgin (\texttt{wes\_NNNNN}) and Camfranglais (\texttt{camf\_NNNNN}). We built it from the labels, which are the same in both varieties, and fixed it across all seeds, conditions, and experiments, so seeds vary only in model initialization and training order. In disjoint-item transfer, fold $k$ trains on source items outside fold $k$ (91--92 items) and tests on target items in fold $k$ (22--23 items), so a test item's counterpart is never in training.}

\medskip
\noindent\textbf{Table G.2: Item numbers in each fold}
\begin{center} \vspace{0.8\baselineskip}
\small
\begin{tabular}{cp{0.8\textwidth}}
\toprule
\textbf{Fold} & \textbf{Item numbers} \\
\midrule
0 & 3, 6, 7, 11, 14, 15, 25, 29, 33, 38, 42, 48, 50, 56, 58, 60, 68, 69, 77, 84, 100, 108, 109 \\
1 & 8, 16, 18, 20, 21, 24, 27, 32, 45, 46, 47, 61, 63, 64, 74, 88, 95, 97, 99, 105, 106, 111, 112 \\
2 & 2, 4, 17, 19, 23, 28, 41, 43, 44, 49, 51, 66, 71, 75, 81, 82, 90, 93, 98, 101, 103, 104, 107 \\
3 & 9, 10, 12, 13, 22, 35, 36, 37, 39, 40, 52, 53, 59, 65, 73, 78, 80, 86, 89, 91, 96, 102, 114 \\
4 & 1, 5, 26, 30, 31, 34, 54, 55, 57, 62, 67, 70, 72, 76, 79, 83, 85, 87, 92, 94, 110, 113 \\
\bottomrule
\end{tabular}
\end{center}

\medskip
\section{Dataset and Audio Checks}
\label{app:dataset_checks}

\subsection{Files and Format}
Each variety is a UTF-8 (with byte-order mark) CSV with 114 rows and seven columns: \texttt{id}, \texttt{premise\_\allowbreak text}, \texttt{premise\_\allowbreak text\_\allowbreak en}, \texttt{premise\_\allowbreak audio}, \texttt{hypothesis\_\allowbreak text}, \texttt{hypothesis\_\allowbreak text\_\allowbreak en}, \texttt{label}. Audio files are referenced as \texttt{audio/\allowbreak pidgin/\allowbreak wes\_\allowbreak NNNNN.wav} and \texttt{audio/\allowbreak camfranglais/\allowbreak camf\_\allowbreak NNNNN.wav}, where \texttt{NNNNN} is the item number (1--114), so item $i$ has the same number in both varieties. 

\noindent Labels are lowercase (entailment, contradiction, neutral), 38 each per variety. No field is empty, no field contains a newline or a curly apostrophe, and all 114 premises and all 114 hypotheses are unique within each variety (114 unique premise--hypothesis pairs).

\subsection{Text Statistics}
\begin{center}\vspace{0.4\baselineskip}
\small
\begin{tabular}{lcc}
\toprule
& \textbf{Pidgin} & \textbf{Camfranglais} \\
\midrule
Items & 114 & 114 \\
Premise words (total / mean / range) & 1,053 / 9.2 / 4--19 & 1,084 / 9.5 / 4--21 \\
Hypothesis words (total / mean / range) & 786 / 6.9 / 3--11 & 845 / 7.4 / 3--15 \\
English-gloss premise words & 1,046 & 1,046 \\
English-gloss hypothesis words & 816 & 816 \\
Premises ending in . ! or ? & 114 & 114 \\
\bottomrule
\end{tabular}
\end{center}
\medskip
\noindent\textit{Word counts are whitespace tokens in the released CSVs. The ASR reference totals in Appendix~\ref{app:asr} (1,055 and 1,096) are slightly higher because normalization turns hyphens and other punctuation into spaces, which splits some tokens (for example, tip-là). No premise contains a question mark. Mean hypothesis length varies little by label (entailment, contradiction,  neutral: Pidgin 6.6, 7.1, 6.9; Camfranglais 7.4, 7.2, 7.6), so length alone is not a visible label cue.}
\medskip

\subsection{Scenario-Matched Structure}
Items are scenario-matched by construction (Section~\ref{sec:sentence}). The Pidgin and Camfranglais sentences differ: the mean word overlap between counterpart premises (shared words divided by all distinct words, per pair) is about 8\%, and no pair is identical. The English glosses are shared across the two files so that items can be aligned, and they are identical in every cell (premise and hypothesis, 114 of 114). Labels agree in all 114 items. For this reason the two English-gloss conditions are treated as one reference condition (Section~\ref{subsec:setup}).

\subsection{Hypothesis Wording Check}

To look for label cues in the hypotheses, we counted hypotheses containing a negation word using simple keyword lists (for example, no, never, nobody for Pidgin and the English glosses; ne, n', pas, jamais, aucun, personne, rien and plus for Camfranglais). This is a heuristic, not a linguistic annotation.

\begin{center}\vspace{0.8\baselineskip}
\small
\begin{tabular}{lccc}
\toprule
\textbf{Hypotheses with a negation word} & \textbf{Entailment} & \textbf{Contradiction} & \textbf{Neutral} \\
\midrule
Pidgin & 6 of 38 (16\%) & 6 of 38 (16\%) & 0 of 38 (0\%) \\
Camfranglais & 10 of 38 (26\%) & 11 of 38 (29\%) & 0 of 38 (0\%) \\
English glosses (shared by both varieties) & 7 of 38 (18\%) & 9 of 38 (24\%) & 0 of 38 (0\%) \\
\bottomrule
\end{tabular}
\end{center}
\medskip
\noindent\textit{No neutral hypothesis contains a negation word, so the presence of negation is a partial label cue. This is consistent with the hypothesis-only results in Section~\ref{subsec:results}, but it does not by itself explain them.}
\smallskip

\subsection{Audio Statistics}
\begin{center}\vspace{0.8\baselineskip}
\small
\begin{tabular}{lcc}
\toprule
& \textbf{Pidgin} & \textbf{Camfranglais} \\
\midrule
Recordings & 114 & 114 \\
Final duration (mean $\pm$ SD, s) & $3.66 \pm 0.94$ & $3.40 \pm 0.79$ \\
Final duration (min / median / max, s) & 2.04 / 3.47 / 6.82 & 2.10 / 3.28 / 5.55 \\
Total duration (s) & 417 & 387 \\
Mean raw clip length before trimming (s) & 5.41 & 5.05 \\
Mean leading silence removed (s) & 1.40 & 1.25 \\
Silence kept after speech (s) & 1.00 (all clips) & 1.00 (all clips) \\
Clips padded with silence to reach 1.0 s & 15 & 9 \\
\bottomrule
\end{tabular}
\end{center}
\smallskip
\noindent\textit{All files are 16 kHz, mono, 16-bit PCM WAV. Speech is located in each file from its own loudness: onset is detected 18 dB below the file's loud-speech level, and each clip starts 0.10 s before the onset and ends 1.0 s after the end of speech (padded with silence if the source ends sooner). Eight clips flagged by an automatic content check were listened to and matched the written text. The preprocessing log (one row per clip, with source file, durations and flags) is included in the release. Durations are shown to two decimals; where the third decimal is exactly 5, the rounding may go either way, so exact values follow: Pidgin minimum 2.045 s, median 3.475 s, maximum 6.815 s, total 416.990 s; Camfranglais minimum 2.105 s, median 3.280 s, maximum 5.555 s, total 387.190 s.}

\end{document}